%% file: main.tex
\documentclass{article}
\usepackage{conference,times}
\input{math_commands.tex}

\usepackage{hyperref}
\usepackage{url}

\usepackage{amsmath} 
\usepackage{graphicx}
\usepackage{amssymb}
\usepackage{wrapfig}
\usepackage{multirow}
\usepackage{makecell}
\usepackage[most]{tcolorbox}
\usepackage{listings}
\usepackage{svg}
\usepackage{wrapfig}
\usepackage{booktabs}
\usepackage[dvipsnames]{xcolor}

\newtcolorbox{promptmint}[1]{
  colframe=green!40!cyan!60!black,
  colback=gray!5!white,
  title=#1,
  fonttitle=\bfseries,
  breakable,
  enhanced,
  sharp corners=south,
  parskip=6pt,
  before upper={\setlength{\parindent}{0pt}\sloppy},
  width=\textwidth
}

\title{ARSM: Auto-Regressive State Machine for Agentic Reasoning Compression}

\author{Xiafeng Man \thanks{Equal Contribution.}\\
College of Future Information Technology\\
Fudan University\\
Shanghai, China \\
\texttt{xfmanacad@outlook.com} \\
\And
Siyuan Ye $^*$ \\
College of Future Information Technology \\
Fudan University \\
Shanghai, China \\
\texttt{syye26@m.fudan.edu.cn} \\
\AND
Xiaosong Ma  \thanks{Corresponding Author.}\\
Department of Computing \\
The Hong Kong Polytechnic University \\
Hong Kong, China \\
\texttt{xiaosong16.ma@connect.polyu.hk} \\
}

\begin{document}

\maketitle

\begin{abstract}
While Large Language Model (LLM)-based agents demonstrate strong capabilities in long-horizon tasks by interleaving reasoning with external environment interactions, the continuous accumulation of context rapidly creates a critical memory bottleneck. Existing memory compression methods rely on task-specific optimization or external auxiliary models, introducing significant computational overhead. Furthermore, the resulting compressed representations tend to lose structured relationships, leading to information dilution, attention collapse, and degraded decision consistency.

To address these limitations, we propose \textbf{A}uto-\textbf{R}egressive \textbf{S}tate \textbf{M}achine (ARSM), a lightweight training-free framework that enables in-situ reasoning compression through structured state evolution. ARSM introduces two key components: (i) a trajectory abstraction mechanism that reorganizes interaction histories into compact \textbf{H}ypothesis--\textbf{A}ction--\textbf{R}esult (HAR) micro-chains; (ii) a dynamic state machine that regulates hierarchical memory through atomic operations and a compression-control parameter. These components are unified within an auto-regressive, self-compressive generation space, where each model output jointly performs external action execution and internal state updates.

We evaluate ARSM on Webshop, Multi-Objective Multi-Hop QA, and SWE-Bench Lite datasets. Experimental results show that ARSM maintains the task performance while simultaneously reducing token consumption, offering a practical, cost‑effective route toward scalable autonomous agents for long‑horizon tasks. 

Code is available at \url{https://github.com/leo-xfm/ARSM}.

\end{abstract}

\section{Introduction}
Current agentic artificial intelligence (Agentic AI) models have shown remarkable capabilities in solving complex tasks. Built upon Chain-of-Thought (CoT) reasoning~\citep{wei2022chain} and tool-use mechanisms~\citep{schick2023toolformer}, these agents typically operate through a multi-turn reasoning–action–observation interaction paradigm~\citep{yao2023react}. In complex domains such as software engineering (SWE), this interleaved reasoning methodology~\citep{yang2024swe} has proven effective in handling long-horizon tasks, including code debugging and error localization.

However, as interaction histories grow, agents must repeatedly process observations that differ substantially in their relevance to future decisions. In software debugging, for example, a file inspection or test run may produce a long output, whereas a later decision may depend on only the hypothesis being tested, the action taken, and the resulting evidence. Retaining every output increases context usage, while removing earlier interactions can discard information needed to interpret subsequent actions. This motivates a memory representation that distinguishes ongoing investigations, completed sub-tasks, and failed attempts, rather than treating all past interactions identically.

To address the challenges above, prior work has explored memory management and context compression strategies. Methods such as LLMLingua~\citep{jiang2023llmlingua} and retrieval augmentation~\citep{fang2025attentionrag} adopt token compression schemes and prune historical contexts. 
Recent approaches~\citep{su2026u, ye2025agentfold} attempt to compress historical trajectories through learned policies, retrieval augmentation, or external distillation modules. 
However, it adds computational overhead and lacks agentic real-time intuition, risking technical detail loss during summarization~\citep{luo2026storage}. 

These observations motivate a more specific question: how can an agent maintain a compact task state that supports subsequent decisions as its interaction history grows? We propose \textbf{A}uto-\textbf{R}egressive \textbf{S}tate \textbf{M}achine (ARSM), a training-free framework for structured context management. ARSM combines compact records of past interactions with explicit rules for updating task progress, failure lessons, milestones, and active reasoning paths. Its central objective is to reduce the amount of historical context carried forward while retaining information relevant to the ongoing task.

Specifically, ARSM consists of two core components.
The first one is the \textbf{trajectory abstraction via \textbf{H}ypothesis--\textbf{A}ction--\textbf{R}esult (HAR) micro-chains}.
We reorganize the conventional Reasoning–-Action–-Observation paradigm into Hypothesis--Action--Observation--Result chain, which can be abstracted into a logical HAR micro-chain. As this is executed by the model itself, it can intrinsically retain the reasoning focus and filter out redundant environmental noise while preserving the underlying structural causal topology of the reasoning process.
The second one is the \textbf{dynamic state machine and sliding window}.
Building upon abstraction, we introduce a dynamic state machine to represent and maintain a hierarchical internal state (e.g., task progress and failure constraints), evolved by a set of atomic operations (e.g., push, fold, prune). 
It also employs a dynamic sliding window mechanism to explicitly bound the token length with a compression-control parameter, allowing for a controllable trade-off between working memory capacity and state sparsity.

\begin{figure}[htbp]
  \centering
  \includegraphics[width=\linewidth, trim=0 2cm 0 2cm]{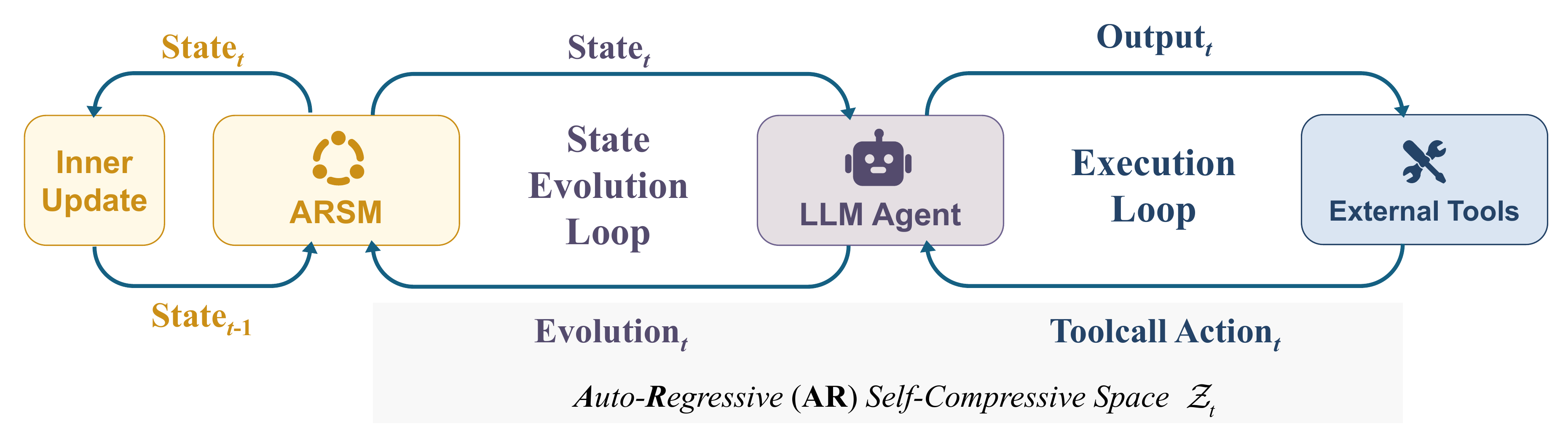}    
  \caption{\textbf{Overview of the dual-loop architecture.} 
    It contains two cyclical processes: (i) execution loop, which manages direct interactions with external tools and feeds back environmental output; and (ii) state evolution loop, which synchronizes agent's cognition with ARSM. The Auto-Regressive Self-Compressive Space acts as a unified generation space, dispatching tool-call action to the environment and evolution metadata to drive ARSM's inner state update: $\text{State}_{t-1} \rightarrow \text{State}_t$.} 
  \label{fig:workflow}
\end{figure}

These two components are then unified within an auto-regressive, self-compressing generation space, where each model output yields: (i) executable actions for environmental interaction, and (ii) structured metadata for updating the internal state. As illustrated in Figure~\ref{fig:workflow}, this establishes a dual-loop architecture that tightly couples external execution with timely internal state evolution. 

We conduct a systematic evaluation of the proposed method on three representative datasets: WebShop~\citep{yao2022webshop}, Multi-Objective Multi-Hop QA~\citep{yang2018hotpotqadatasetdiverseexplainable, xanh2020_2wikimultihop, trivedi2021musique, aksitov2023restmeetsreactselfimprovement}, and SWE-Bench Lite~\citep{jimenez2024swebench}. 
Experimental results show that on WebShop, ARSM consistently improves the reward across different backbone model families; on 16-Objective Multi-Hop QA, it achieves improved EM and F1 scores. Across all evaluated datasets, ARSM substantially reduces token costs by up to 82\%.
These findings demonstrate that structured chain construction and state compression are key factors in improving both efficiency and performance in long-horizon reasoning.

Our main contributions are summarized as follows:

\begin{itemize}
    \item We introduce \textbf{A}uto-\textbf{R}egressive \textbf{S}tate \textbf{M}achine (ARSM), a training-free context-management framework in which the agent generates memory-update metadata alongside its next action, without requiring fine-tuning or an auxiliary compression model.
    
    \item We combine Hypothesis–Action–Result records with an event-driven controller that manages task progress, failure lessons, milestones, and active interaction records. A compression control parameter adjusts the capacities of structured memory and recent raw history.

    \item We evaluate ARSM on WebShop, Multi-Objective Multi-Hop QA, and SWE-bench Lite. The results characterize its task-dependent benefits and cost–accuracy trade-offs, while component comparisons and trajectory analyses examine the behavior and limitations of the proposed memory design.
\end{itemize}

\section{Related Work}

\paragraph{Interleaved Reasoning in Agents.} 
Building upon Chain-of-Thought (CoT) reasoning~\citep{wei2022chain} and tool-use integration~\citep{schick2023toolformer, chen2024agent}, LLM-based agents employing interleaved reasoning paradigms, characterized by multi-turn interactions~\citep{yao2023react, shinn2023reflexion}, have demonstrated superior performance in solving complex problems. 
Recent work~\citep{jin2025search, chen2024agent} has further improved general-purpose agentic capabilities through task-specific adaptation.
In specialized domains such as software engineering (SWE), frameworks such as SWE-Agent~\citep{yang2024swe} and SE-Agent~\citep{guo2025se} deploy professional tool functions and reuse interaction trajectories to construct evolving knowledge bases. 
Despite these advances, most existing approaches still rely on the monotonic accumulation of raw context, inevitably suffering from information dilution and attention degradation as the interaction horizon expands~\citep{li2023unlocking}.

\paragraph{Agent Memory and Context Compression.} 
To support long-term reasoning, recent research has increasingly focused on memory management and context compression, integrating retrieval-augmented generation (RAG)~\citep{lewis2020retrieval, jin2025search} to integrate external knowledge. 
For memory management, some agents~\citep{park2023generative, zhong2024memorybank} utilize persistent memory stores to simulate long-term interactions, while systems like MemGPT~\citep{packer2023memgpt} and LONGMEM~\citep{wang2023augmenting} enhance LLMs with scalable, hierarchical memory architectures. 
To directly mitigate sequence length, compression methods~\citep{jiang2023llmlingua, li2023compressing, wang2026swe} prune redundant information at the token level, whereas attention-based approaches~\citep{fang2025attentionrag} optimize efficiency through importance-guided selection. 
Within agent frameworks, most approaches~\citep{kang2025acon} employ a tiny auxiliary LLM to dynamically distill or truncate trajectories; others, like MEM1~\citep{zhou2025mem1}, coordinate between memory selection and the active reasoning process. ContextBudget~\citep{wu2026contextbudgetbudgetawarecontextmanagement} formulates context management as a budget-constrained sequential decision problem and learns when and how to compress interaction history through reinforcement learning. 
Recent studies~\citep{su2026u, ye2025agentfold} have also proposed trajectory folding strategies to truncate growing contexts. 
However, most of these approaches depend on external auxiliary modules or separate processing stages. It not only introduces significant computational overhead, but fundamentally disconnects the compression mechanism from primary agent's active reasoning state. 

\section{Methodology}

We propose \textbf{A}uto-\textbf{R}egressive \textbf{S}tate \textbf{M}achine (ARSM), a lightweight training-free architecture that manages context through a repeated cycle of observation, reflection, state update, and action. After receiving the outcome of a previously executed action, the agent generates a reflection on that outcome, metadata for updating its memory, and a proposal for the next action. The controller uses this metadata to update the structured state and applies the configured retention limits when assembling subsequent context. Section~\ref{sec: traj abstract} describes the output format and HAR records, Section~\ref{sec: state machine} defines the memory components and their event-conditioned updates, and Section~\ref{sec: sliding_window} explains how the compression-control parameter adjusts memory capacities.

\subsection{Auto-Regressive Chain-Structured Trajectory Abstraction}
\label{sec: traj abstract}

\begin{figure}[htbp]
  \centering
  \includegraphics[width=0.9\linewidth, trim=0 0 0 0]{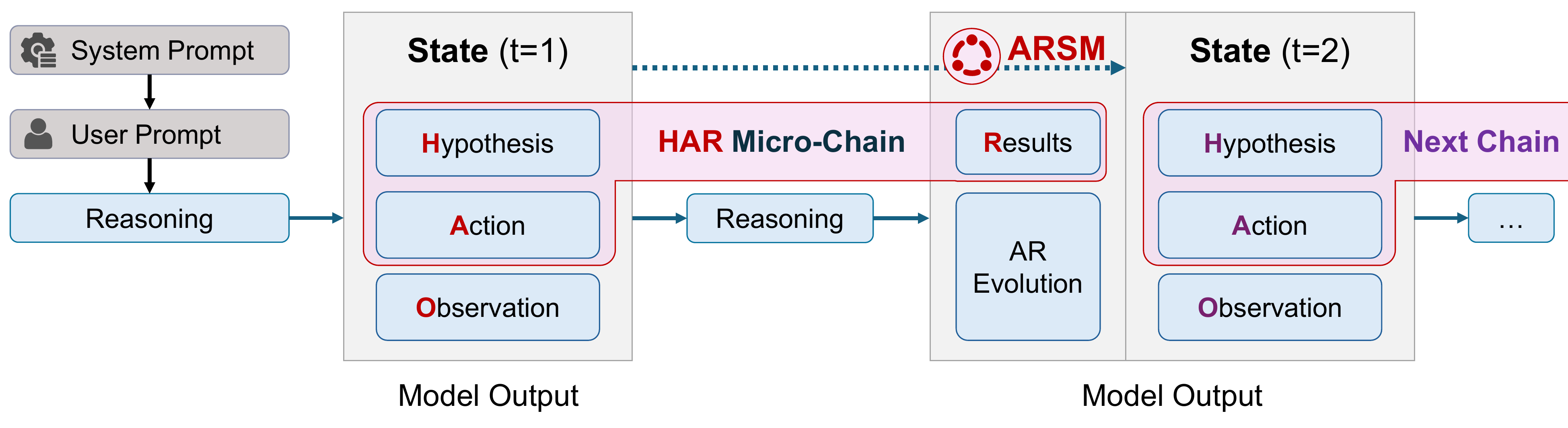}
    \caption{\textbf{Workflow of trajectory abstraction.} The agent performs \textit{in-situ} compression: previous raw observations are evaluated and distilled into concise results.}  
    \label{fig:chain}
\end{figure}

\paragraph{Auto-Regressive (AR) Self-Compressive Space}
As shown in Figure~\ref{fig:chain}, we define most models' responses as a structured space $\mathcal{Z}_t$, which falls into three components: conclusions from the previous turn, the direction for the next turn, and the specific action to be executed. 
Specifically, it is formatted as explicit, structured metadata (e.g., progress reflections, status updates, pitfall extractions) auto-regressively, aligned with its standard reasoning steps.
Upon the completion of each reasoning turn, the space $\mathcal{Z}_t$ is structurally parsed as $\mathcal{Z}_t \mapsto \langle \tau_t, \mathbf{u}_t \rangle$, 
where $\tau_t$ represents the extracted causal tuple with precise interaction details, such as reflections and next-turn aims, 
and $\mathbf{u}_t \in \{0, 1\}^{4 \times 1}$ represents a discrete one-hot routing vector. 
By mapping the generated status updates into four distinct operational branches (\textit{Ongoing, Milestone, Failure, Success}), $\mathbf{u}_t$ serves as a routing signal to direct the transition mechanics within our state machine.

\paragraph{Micro Hypothesis--Action--Result (HAR) Chain}
The traditional agent trajectory $\mathcal{T}$ mainly contains three sequential parts: reasoning, action, and output. To strengthen the causal link in reasoning, we separate the generated message and action of the assistant as a detailed sequence of interactions, apart from reasoning:
\begin{equation}
\mathcal{T}_t = \{ (R_i, H_{i+1}, A_{i+1}, O_{i+1}) \}_{i=1}^{t-1}
\end{equation}
where \textbf{Hypothesis \& Aim} ($H_i$) is derived from the forward-looking portion of the messages and defines the objective for the next action; \textbf{Action} ($A_i$) represents the exact tool invocation command; \textbf{Raw Observation} ($O_i$) contains the tool's raw execution output, which is mostly redundant; and \textbf{Result \& Reflection} ($R_i$) is a concise distillation of $O_i$ that provides insights for the next turn. Specifically, the raw observation $O_i$ frequently contains thousands of tokens of logs, which inevitably obscures the crucial causal links between past reasoning and future actions. To resolve this, our framework compresses the raw trajectory into a highly sparsified HAR chain (see Figure~\ref{fig:chain}):
\begin{equation}
\tau_t = (H_t, A_t, R_t)
\end{equation}
where such chain abstraction is designed to preserve the causal topology of the debugging process while substantially reducing the agent from the token burden of raw environmental feedback.

\subsection{Dynamic State Machine}
\label{sec: state machine}

The core of our framework is a dynamic state machine that manages the continuous evolution of the agent's internal cognitive state. As illustrated in Figure~\ref{fig:workflow}, the system state $\mathcal{S}_t$ evolves through a formal state evolution loop driven by the agent's self-generated metadata:
\begin{equation}
\label{equ: state_trans}
\mathcal{S}_t = \Delta_{\mathcal{Z}_{t}} (\mathcal{S}_{t-1}) \in \mathbb{S}^{n \times 1}
\end{equation}
where $n$ is the number of heterogeneous internal state components, and $\Delta$ denotes the core transition operator of our dynamic state machine parameterized by the structured metadata space $\mathcal{Z}_t$. 
Following this update, the newly generated state $\mathcal{S}_t$ with the raw observation is fed back into the agentic model to guide the subsequent reasoning step.

\paragraph{Internal State Definition}
In our settings, we define the system $\mathcal{S}_t$  as:
\begin{equation}
\mathcal{S}_t = [c_t, p_t, w_t, m_t, a_t]^T
\end{equation}
where its elements are defined as follows: (i) \textbf{Control Layer ($c_t$)} acts as a global checklist for macroscopic task decomposition and progress tracking; (ii) \textbf{Constraint Layer ($p_t, w_t$)} utilizes pitfall memory ($p_t$) to prevent redundant failures and loop alerts ($w_t$) to intervene during circular reasoning; (iii) \textbf{Knowledge Layer ($m_t$)} stores high-level logical milestones that fold completed sub-tasks of macro-chains into verified facts in order to free context; and (iv) \textbf{Working Memory Layer ($a_t$)} maintains the active causal reasoning trajectory of the current micro-chain $\{ \tau_i \}$ as compressed HAR tuples, retaining immediate causal context without the overhead of raw logs.

\paragraph{State Transition Matrix}

Operating alongside the execution loop, the state evolution is governed by the core transition operator $\Delta$. Directed by the previously parsed routing vector $\mathbf{u}_t$, the transition acts as a gate to selectively activate the corresponding operational branch. The unified state transition $\Delta$ in Equation~\ref{equ: state_trans} is mathematically expanded as a matrix projection:
\begin{equation}
\mathcal{S}_t 
= \Delta_{\mathcal{Z}_t}(\mathcal{S}_{t-1}) 
= \Big( \mathbf{T}(\tau_t) \circledast \mathcal{S}_{t-1} \Big) \mathbf{u}_t
\end{equation}
where $\mathbf{T}(\tau_t) \in \mathbb{O}^{5 \times 4}$ is the \textbf{transition operator matrix} instantiated by the tuple $\tau_t$, and $\circledast$ denotes the \textbf{element-wise operator}. It broadcasts the state vector $\mathcal{S}_{t-1}$ into the operator matrix space, such that each atomic operator is applied to its corresponding layer, generating a candidate state matrix. Specifically, $\mathbf{T}(\tau_t)$, constructed of atomic operators (detailed in Appendix~\ref{appendix: atomic operator}), is defined as:
\begin{equation}
\mathbf{T}(\tau_t) = \begin{bmatrix}
\mathbb{I} & \mathbb{I} & \chi(\cdot) & \mathbb{I} & \cdot \oplus \tau_t \\ 
\delta(\cdot) & \mathbb{I} & \emptyset & \cdot \cup \phi(\tau_t) & \emptyset \\ 
\mathbb{I} & \cdot \cup \ell(\tau_t) & \mu(\cdot, \tau_t) & \mathbb{I} & \pi(\cdot) \\ 
\mathbb{I} & \mathbb{I} & \chi(\cdot) & \mathbb{I} & \cdot \oplus \tau_t 
\end{bmatrix}^T
\end{equation}
where $\mathbb{I}$ denotes the identity operator that maintains the previous state, and $\emptyset$ denotes a selective truncation operator that discards certain states before milestone $\tau_t$ while retaining recent activations.

\begin{figure}[htbp]
  \centering
  \includegraphics[width=\linewidth,  trim=0 1cm 0 1cm]{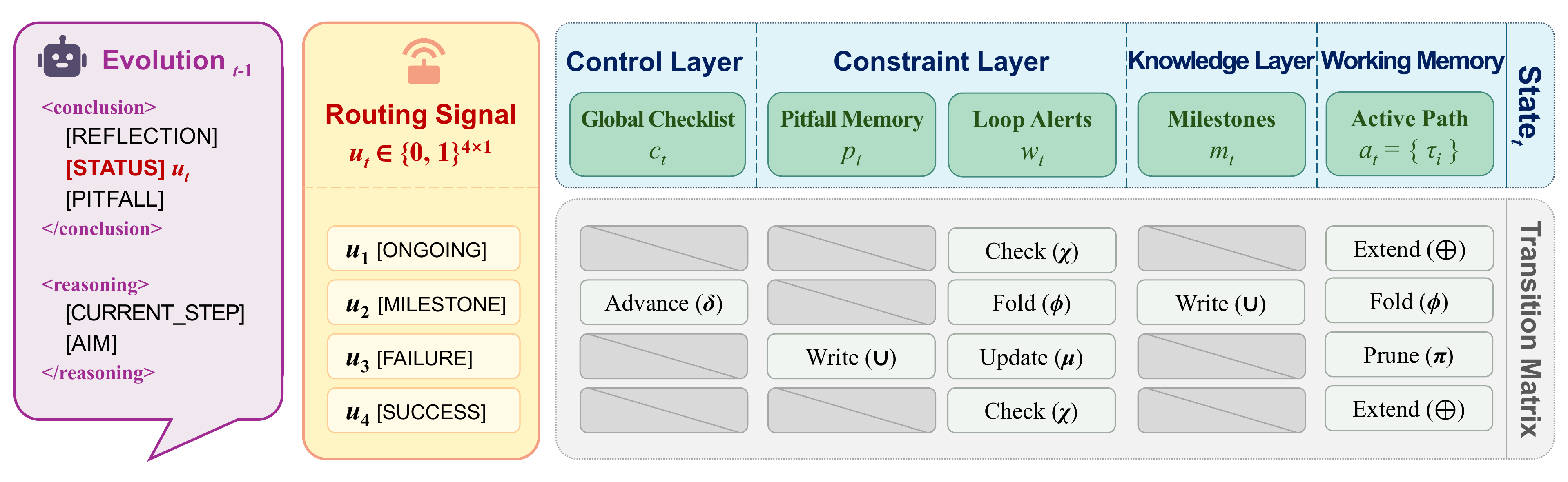}
  \caption{\textbf{State transition mechanism.} The model's structured generation space is parsed to yield a discrete routing signal $u_t$, mapping to the branches. Targeted atomic operators are then dynamically applied to update their respective layers, while unaffected subspaces remain frozen.}  
    \label{fig:State Transition}
\end{figure}

With the element-wise broadcasting of $\mathcal{S}_{t-1}$ across the matrix, the gating vector $\mathbf{u}_t$ activates a specific evolutionary branch identified by the judgment block: 
\begin{itemize}
    \item \textbf{Ongoing} ($u_1=1$): When the model continues to solve a sub-task without yet reaching a definitive conclusion, control and constraint layers remain frozen ($\mathbb{I}$), the active path extends ($\oplus$) with the latest interaction, and loop alerts are checked ($\chi$); 
    \item \textbf{Milestone} ($u_2=1$): Upon resolving a sub-task, ARSM summarizes acquired knowledge by first folding active paths ($\phi$) into the knowledge layer and truncating them ($\emptyset$), and then advancing the checklist ($\delta$) and clearing loop alerts ($\emptyset$); 
    \item \textbf{Failure} ($u_3=1$): When encountering execution errors or dead ends, to penalize the flawed trajectory and prevent the repetition of identical mistakes, the model extracts failure reasons ($\ell$) and updates loop alerts ($\mu$). Invalid active path is then pruned ($\pi$) to its last verified node;
    \item \textbf{Success} ($u_4=1$): It extends ($\oplus$) with the confirmed result and performs a final loop alert check ($\chi$), marking a valid and successful execution. 
\end{itemize}
Through these mechanisms, ARSM effectively mitigates context bloat and logical contamination, thereby sustaining efficiency and consistency across complex, long-horizon reasoning tasks.

\subsection{Sparsified Context Management}
\label{sec: sliding_window}

ARSM transforms the updated state $\mathcal{S}_t$ into a sparsified context window systematically partitioned into three blocks: (i) \textbf{Structured System State} stores compressed context to ensure goal alignment without token bloat; (ii) \textbf{Episodic Raw Trajectory} retains recent high-fidelity interactions for execution tracking; and (iii) \textbf{Operational Directives} provides explicit instructions to enforce the mandated output schema.
To dynamically regulate these block sizes while preserving reasoning continuity, we introduce a continuous compression-control parameter $\rho \in [0, 1]$. The dynamic window size $W(\rho)$ for a given memory component is defined as:
\begin{equation}
    \label{equ: window}
    W(\rho) = 
    \begin{cases} 
    L_{\max} & \rho = 0 \\ 
    \max \left( V_{\min}, \min \left( L_{\max}, V_{\min} + \left\lfloor (V_r - V_{\min}) \cdot \frac{1-\rho}{\rho} \right\rceil \right) \right) & \rho > 0 
    \end{cases}
\end{equation}
where $V_r$ is the target capacity at $\rho=0.5$, $V_{\min}$ sets the minimum operational fidelity floor for the component, $L_{\max}$ denotes the global system context limit, and $\lfloor \cdot \rceil$ denotes rounding to the nearest integer. More details are listed in Appendix~\ref{appendix: state_compression_ratio}.

\section{Experiments}

\subsection{Datasets and Baselines} 
We evaluate ARSM on three datasets: WebShop~\citep{yao2022webshop}, a simulated e-commerce website environment with real-world products and crowd-sourced text instructions; Multi-Objective Multi-Hop QA~\citep{yang2018hotpotqadatasetdiverseexplainable, xanh2020_2wikimultihop, trivedi2021musique, aksitov2023restmeetsreactselfimprovement} following MEM1~\citep{zhou2025mem1}; and SWE-Bench Lite~\citep{jimenez2024swebench}, which comprises 300 GitHub repository issues. 
These settings cover software issue debugging, web navigation, tool interaction, and multi-hop reasoning. We employ Qwen2.5-7B-Instruct, Qwen2.5-14B-Instruct~\citep{qwen2.5, qwen2}, Qwen3.5-9B, Qwen3.5-27B~\citep{qwen3.5}, DeepSeek-V4.1-Flash~\citep{deepseekai2026deepseekv41flash}, and
GLM-5.3-Flash~\citep{glm5team2026glm5vibecodingagentic} as our backbone models. 
Detailed experimental settings and evaluation metrics are illustrated in Appendix~\ref{appendix: exp_setup} and ~\ref{appendix: eval_metrics}.

\subsection{Compared Methods}

We compare ARSM against several baselines:
Search-R1~\citep{jin2025search}, SWE-Agent~\citep{yang2024swe}, SE-Agent~\citep{guo2025se}, and Agent-FLAN~\citep{chen2024agent} are reasoning-driven methods; SWE-Pruner~\citep{wang2026swe}, MEM1~\citep{zhou2025mem1}, and ContextBudget~\citep{wu2026contextbudgetbudgetawarecontextmanagement} are memory-aware methods for efficient reasoning.

\begin{table}[htbp]
    \caption{\textbf{Comparison of training costs, structural dependencies, and generalizability across methods.} ``$\circ$'' indicates that the generality requires task-specific training.}
    \label{table:method_comprehensive_comparison}
    \centering
    \setlength{\tabcolsep}{5pt}
    \begin{tabular}{lccccc}
        \toprule
        Method & Training-Free & Self-Contained & Plug-and-Play & Generality \\
        \midrule
        Search-R1               & $\times$~(RL)   & \checkmark       & $\times$     & $\times$~(QA) \\
        MEM1          & $\times$~(RL)   & $\times$~(RAG)      & $\times$     & $\circ$ \\
        ContextBudget           & $\times$~(RL)   & \checkmark      & $\times$     & $\circ$ \\
        Agent-FLAN              & $\times$~(SFT)  & \checkmark        & $\times$     & $\circ$ \\
        Mini SWE-Agent          & \checkmark      & \checkmark        & \checkmark   & $\times$~(Code) \\
        SE-Agent                & \checkmark      & \checkmark       & \checkmark   & $\times$~(Code) \\
        SWE-Pruner              & \checkmark      & $\times$~(Auxiliary)     & $\times$     & $\times$~(Code) \\
        \midrule
        ARSM (Ours)    & \checkmark & \checkmark  & \checkmark & \checkmark \\
        \bottomrule
    \end{tabular}
\end{table}

As summarized in Table~\ref{table:method_comprehensive_comparison}, ARSM operates under a lightweight regime, whereas several baselines rely on fine-tuning, reinforcement learning, retrieval augmentation, or auxiliary models, giving them access to additional optimization or inference capacity. 
Since no prior method operates under exactly the same regime as ARSM, we include them as comparative reference points, making the comparison conservative for ARSM.

\subsection{Main Results}

In this section, we first examine whether ARSM improves the balance between task quality and token usage in each evaluation setting. We report these outcomes jointly, since lower token usage may accompany either improved performance or reduced accuracy. Section~\ref{sec: ablation} then examines the behavior of the memory controller, compares simplified configurations, and studies sensitivity to the compression-control parameter.

\paragraph{WebShop} 
As shown in Table~\ref{table: result_webshop}, ARSM improves task reward while substantially reducing token consumption, not confined to a single backbone family: it appears on Qwen, DeepSeek, and GLM.

\begin{table}[htbp]
	\caption{ Performance on WebShop.}
	\label{table: result_webshop}
	\centering
	\begin{tabular}{lccc}
		\toprule
		Method 	& Reward $\uparrow$ & Peak Token (k) $\downarrow$ & Total Token (k) $\downarrow$ \\
		\midrule
		Qwen2.5-7B-Instruct        & 0.2858   & 5.41 ± 2.40 & 63.24 \\ 
		Agent-FLAN-7B (SFT)   & 0.2342    & 5.84 ± 1.61  {\color{Red}($\uparrow$ 7.9\%)} & 63.26 {\color{Red}($\uparrow$ 0.03\%)} \\
		MEM1-Webshop (Qwen, RAG+RL)       & 0.1528   & 1.55 ± 0.14 {\color{ForestGreen}($\downarrow$ 71\%)}   & 21.82 {\color{ForestGreen}($\downarrow$ 65\%)} \\
		ARSM (Qwen-7B, $\rho=0.3$) & 0.4906 &  3.61 ± 0.87 {\color{ForestGreen}($\downarrow$ 33\%)} &  21.46 {\color{ForestGreen}($\downarrow$ 66\%)}\\
		ARSM (Qwen-7B, $\rho=0.5$) & 0.4641 &  3.01 ± 0.47 {\color{ForestGreen}($\downarrow$ 44\%)} &  23.11 {\color{ForestGreen}($\downarrow$ 63\%)}\\
		ARSM (Qwen-7B, $\rho=0.7$) & 0.4186 &  2.51 ± 0.23 {\color{ForestGreen}($\downarrow$ 54\%)} &  20.88 {\color{ForestGreen}($\downarrow$ 67\%)}\\
		\midrule
		Qwen2.5-14B-Instruct       & 0.4464   & 5.46 ± 4.23 & 45.34 \\ 
		ARSM (Qwen-14B, $\rho=0.3$) & 0.5881 & 3.84 ± 1.66  {\color{ForestGreen}($\downarrow$ 30\%)} & 24.05 {\color{ForestGreen}($\downarrow$ 47\%)}\\
		ARSM (Qwen-14B, $\rho=0.5$) & 0.5736 & 3.23 ± 1.05  {\color{ForestGreen}($\downarrow$ 41\%)} & 21.01 {\color{ForestGreen}($\downarrow$ 54\%)}\\
		ARSM (Qwen-14B, $\rho=0.7$) & 0.5582 & 2.70 ± 0.62  {\color{ForestGreen}($\downarrow$ 51\%)} & 20.77 {\color{ForestGreen}($\downarrow$ 54\%)}\\
		\midrule
		Deepseek-V4.1-Flash    & 0.6151 & 18.24   &  33.90 \\ 
		ARSM (Deepseek, $\rho=0.5$) & 0.6633 & 11.89 {\color{ForestGreen}($\downarrow$ 35\%)} & 20.04 {\color{ForestGreen}($\downarrow$ 41\%)}\\
		\midrule
		GLM-5.3-Flash    & 0.6441 & 28.53   &  37.15 \\ 
		ARSM (GLM, $\rho=0.5$) & 0.7025 & 11.33 {\color{ForestGreen}($\downarrow$ 60\%)} & 24.25 {\color{ForestGreen}($\downarrow$ 35\%)}\\
		\bottomrule
	\end{tabular}
\end{table}

\paragraph{Multi-Objective Multi-Hop QA (16-Objective)} 
As shown in Table~\ref{table: result_16obj}, ARSM substantially improves both $\Sigma_{\mathrm{EM}}$ and $\Sigma_{\mathrm{F1}}$ (sums over 16 objectives) while maintaining compact context. Across different compression-control parameters, ARSM consistently outperforms other methods.

\begin{table}[htb]
	\caption{\textbf{Performance on Multi-Objective Multi-Hop QA (16-Objective).} N/R denotes results that were not reported in the original work.}
	\label{table: result_16obj}
	\centering
	\begin{tabular}{lcccc}
		\toprule
		Method 	& $\Sigma_{\mathrm{EM}}$ $\uparrow$ & $\Sigma_{\mathrm{F1}}$ $\uparrow$ & Peak Token (k) $\downarrow$ & Total Token (k) $\downarrow$ \\
		\midrule
		Qwen2.5-14B-Instruct       & 0.567   & 0.703  &  3.84 & N/R \\
		Search-R1 (RL)             & 0.009  
		& 0.011 & 2.09  {\color{ForestGreen}($\downarrow$ 46\%)} & N/R \\
		MEM1-QA (14B, truncate)       & 0.44
		& 0.50  & 1.19  {\color{ForestGreen}($\downarrow$ 69\%)} & N/R \\
		ContextBudget\footnotemark (30B, w/o RL) & 0.17 
		& N/R & 8.00 {\color{red}($\uparrow$ 108\%)} & N/R \\
		\midrule
		ARSM (14B, $\rho=0.3$) & 1.85 & 2.15 
        & 0.78 {\color{ForestGreen}($\downarrow$ 80\%)} & 55.06\\
		ARSM (14B, $\rho=0.5$) & 1.87 & 2.30
		& 0.81  {\color{ForestGreen}($\downarrow$ 79\%)} & 54.09\\
		ARSM (14B, $\rho=0.7$) & 2.01 & 2.34
        & \underline{0.79} {\color{ForestGreen}($\downarrow$ 79\%)} & 51.43\\
		\bottomrule
	\end{tabular}
\end{table}

\footnotetext{ContextBudget reports only the context budget; therefore, we use it as its peak token usage.}

\begin{figure}[htbp]
	\centering
	\includegraphics[width=\linewidth]{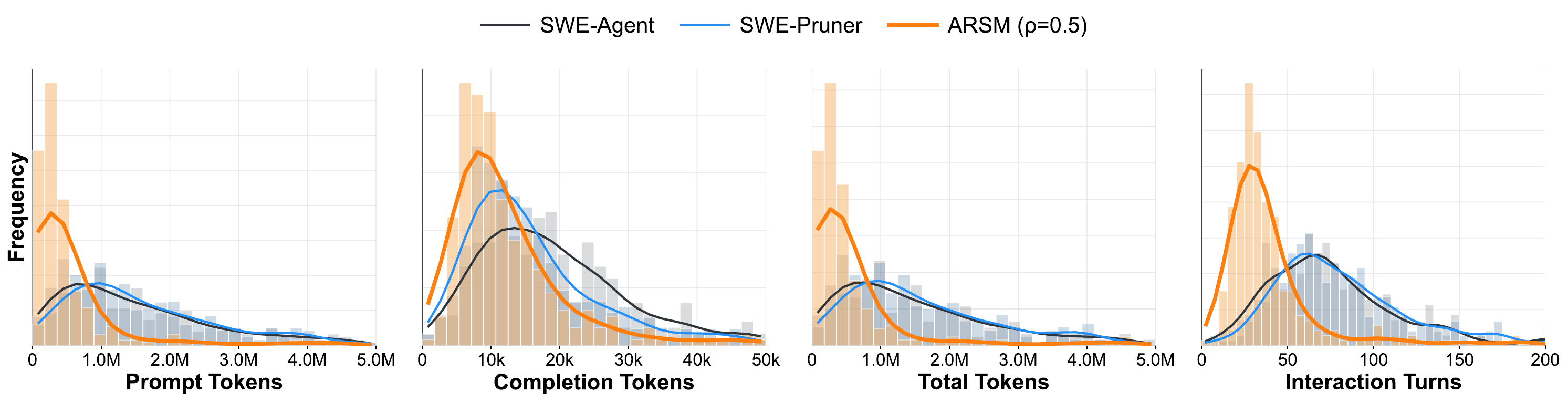}
	\caption{\textbf{Efficiency analysis on SWE-Bench Lite using Qwen3.5-27B.} It illustrates the distribution of tokens and interaction turns. Our ARSM ($\rho=0.5$) consistently shifts the distribution towards the left, demonstrating higher efficiency compared to baselines.}
	\label{fig: turn_27b}
\end{figure}

\paragraph{SWE-Bench Lite}
As shown in Table~\ref{table: result_swe}, ARSM consistently establishes a controllable performance, efficiency trade-off across different model scales. We also analyze the token consumption and interaction turns in Figure~\ref{fig: turn_27b}, indicating reduced prompt payload and accelerated reasoning.

\begin{table}[htbp]
  \caption{Performance on SWE-Bench Lite.} 
  \label{table: result_swe}
  \centering
  \begin{tabular}{lccccc}
    \toprule
    \multirow{2}{*}{Method} & \multicolumn{3}{c}{Qwen3.5-9B} & \multicolumn{2}{c}{Qwen3.5-27B} \\ 
    \cmidrule(lr){2-4} \cmidrule(lr){5-6}
    & Pass@1 $\uparrow$ & Pass@5 $\uparrow$ & Token (M) $\downarrow$ & Pass@1 $\uparrow$ & Token (M) $\downarrow$\\
    \midrule
    Mini SWE-Agent        & 26.6\%   & 47.5\%
                            & 6.57
                            & 51.7\%
                            & 1.74 \\
    SE-Agent\footnotemark (Evolution) 
                         & 15.0\%   & N/A & N/A & N/A & N/A \\
    SWE-Pruner (Auxiliary)           & 40.7\%    & 58.7\%
                        & 2.29 {\color{ForestGreen}($\downarrow$ 65\%)}
                        & 45.0\% 
                        & 2.37 {\color{red}($\uparrow$ 36\%)}\\
    \midrule
    ARSM ($\rho=0.1$) & 29.7\% & 46.67\% 
                                & 2.40 {\color{ForestGreen}($\downarrow$ 63\%)}
                                & 46.5\%
                                & 0.92 {\color{ForestGreen}($\downarrow$ 47\%)}\\
    ARSM ($\rho=0.5$) & 26.2\% & 44.5\%
                                & 2.05 {\color{ForestGreen}($\downarrow$ 69\%)}
                                & 46.0\%
                                & 0.84 {\color{ForestGreen}($\downarrow$ 52\%)}\\
    ARSM ($\rho=0.9$) & 20.3\% & 53.3\%  
                                & 1.15 {\color{ForestGreen}($\downarrow$ 82\%)}
                                & 38.0\% 
                                & 0.47 {\color{ForestGreen}($\downarrow$ 73\%)}\\
    \bottomrule
  \end{tabular}
\end{table}

\footnotetext{We were unable to complete the evaluation for SE-Agent; its complex system prompts and tool-calling requirements overwhelmed the models, consistently resulting in format errors or empty patches. }

\subsection{Ablation Studies}
\label{sec: ablation}

\begin{figure}[htbp]
    \centering
    \includegraphics[width=\linewidth]{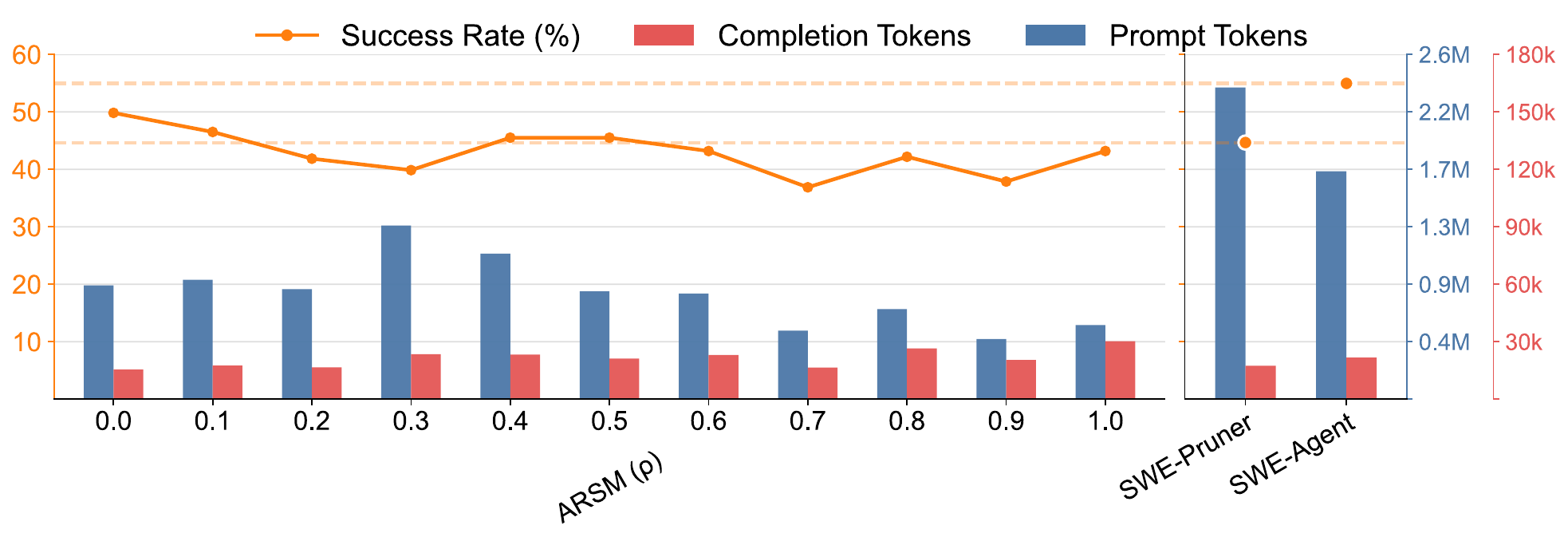}
    \caption{\textbf{Performance and efficiency comparison across different ARSM compression-control parameter ($\rho$) and baselines using Qwen3.5-27B.} ARSM drastically reduces prompt token usage compared to SWE-Agent and SWE-Pruner across all evaluated compression rates.}
    \label{fig: ablation_27b}
\end{figure}

\paragraph{Compression Control}
To investigate the sensitivity of our core mechanism, we conduct an ablation study on the compression-control parameter $\rho$. Figure~\ref{fig: ablation_27b} shows that $\rho$ governs the trade-off between the retention of contextual details and the aggressiveness of token pruning.

\paragraph{Routing Behavior Analysis.}
As shown in Table~\ref{tab:routing_distribution}, the routing behavior of ARSM ($\rho=0.5$) indicates that it actively invokes milestone consolidation, failure handling, and completion transitions rather than collapsing into an append-only running memory.

\begin{table}[htbp]
    \centering
    \caption{\textbf{Distribution of routing signals on SWE-Bench Lite, Qwen3.5-9B.}}
    \label{tab:routing_distribution}
    \begin{tabular}{ccccc}
        \toprule
        Routing Signal 
        & \textsc{Ongoing} 
        & \textsc{Milestone} 
        & \textsc{Failure} 
        & \textsc{Success} \\
        \midrule
        Percentage (\%) 
        & 76.34 
        & 8.99 
        & 9.17 
        & 5.50 \\
        \bottomrule
    \end{tabular}
\end{table}

\paragraph{Trajectory Summarization and State Machine Evolution}
We compare ARSM with several simplified variants on Qwen2.5-7B-Instruct, as shown in Table~\ref{tab:state_summary_ablation}. 
Structured summary compresses aggressively but suffers substantial performance degradation. HAR-only mostly improves task performance, while SM-only mainly reduces memory cost with limited performance change. Combining both yields the strongest performance--efficiency trade-off.
More details are provided in Appendix~\ref{appendix: exp_protocol}.

\begin{table}[htbp]
    \centering
    \caption{\textbf{Trajectory summarization and state machine (SM) evolution on 200 Webshop instances.} It highlights the complementary roles of HAR abstraction and dynamic state evolution in balancing task performance and token efficiency.}
    \label{tab:state_summary_ablation}
    \begin{tabular}{lccccc}
        \toprule
        Method 
        & HAR & SM & Reward $\uparrow$  & Peak Tokens (k) $\downarrow$
        & Total Tokens (k) $\downarrow$ \\
        \midrule
        Plain Summary  & $\times$ & $\times$
        & 0.3305 
        & 3.63 ± 0.84 
        & 44.71 \\

        Structured Summary  & $\times$ & $\times$
        & 0.1321
        & 1.37 ± 0.10
        & 18.44 \\
        
        HAR-only & \checkmark & $\times$
        & 0.4240
        & 3.18 ± 1.86
        & 28.92 \\

        SM-only & $\times$ & \checkmark
        & 0.3178
        & 2.74 ± 0.23
        & 24.18 \\
        
        ARSM ($\rho=0.5$) & \checkmark & \checkmark 
        & 0.4899 &  2.97 ± 0.45 &  22.84 \\
        \bottomrule
    \end{tabular}
\end{table}

\paragraph{Trajectory-Level Evaluation.}
We evaluate ARSM ($\rho=0.5$) using an LLM-based evaluator. As Table~\ref{tab:webshop_state_quality} shows, ARSM preserves factual fidelity and decisive information while maintaining routing and trajectory coherence. Details are provided in Appendix~\ref{appendix: exp_protocol}.

\begin{table}[htbp]
    \centering
    \caption{
        \textbf{Trajectory-level evaluation of ARSM state updates and routing
        decisions on 100 WebShop trajectories with $\rho=0.5$.}
        Higher scores indicate better performance.
    }
    \label{tab:webshop_state_quality}
    \begin{tabular}{lcc}
        \toprule
        Metric & Score $\uparrow$ & Interpretation \\
        \midrule
        State Fidelity       & 76.1\% & Factual faithfulness of compressed states \\
        State Sufficiency    & 81.4\% & Retention of information needed later \\
        Routing Correctness  & 73.2\% & Correctness of routing signals \\
        Trajectory Coherence & 72.5\% & Global consistency across the trajectory \\
        \bottomrule
    \end{tabular}
\end{table}

\paragraph{More Experiments.}
We also conduct the following studies, and those details are provided in Appendix~\ref{appendix: more_experiment},~\ref{appendix:state_representation}, and~\ref{appendix:rho_case_study}.

\begin{itemize}
    \item \textbf{Efficiency and compression-control analysis on Qwen3.5-9B, SWE-Bench Lite.} ARSM reduces token and turn costs, while $\rho$ controls the retention--compression trade-off.
    \item \textbf{Solved-set overlap.} We examine solved instances across different $\rho$ settings, revealing a stable solving core and complementary coverage across compression regimes.    
    \item \textbf{State representation analysis.} We illustrate the two complementary memory surfaces of ARSM: recent raw interaction memory and compressed internal state, showing how they respectively preserve local interaction fidelity and long-range reasoning continuity.
    \item \textbf{Compression-control parameter case study.} Using the same instance under different $\rho$ settings, we qualitatively examine how increasing compression shifts task-relevant information from transcript-level raw interaction history toward structured internal state.
\end{itemize}

\section{Conclusion}

In this work, we identify monotonic accumulation of raw context as a fundamental bottleneck for LLM-based agents in interleaved reasoning.
To overcome the resulting attention sink and computational bloat, we introduce the \textbf{A}uto-\textbf{R}egressive \textbf{S}tate \textbf{M}achine (ARSM), a lightweight, self-contained framework for in-situ context compression. 
By abstracting raw interactions into \textbf{H}ypothesis-\textbf{A}ction-\textbf{R}esult (HAR) micro-chains and orchestrating them through a dynamic state machine, ARSM shifts the agentic paradigm from passive log accumulation to structured state evolution.

Our systematic evaluation across WebShop, Multi-Objective Multi-Hop QA, and SWE-Bench Lite demonstrates that ARSM can substantially reduce context and token costs while improving strong long-horizon reasoning performance in most evaluated settings. 
Crucially, these efficiency gains are obtained without training or auxiliary compression models, relying instead on the backbone model's intrinsic reasoning and summarization capabilities.
Furthermore, by formalizing a compression-control parameter, ARSM provides a tunable mechanism to regulate the context, empowering engineers to explicitly balance operational fidelity against inference cost based on task demands and model capacities.

Overall, our findings suggest that structured state evolution is a promising alternative to continuously expanding interaction histories for long-horizon agents. By autonomously distilling, folding, and pruning working memory, ARSM provides a scalable and cost-effective foundation for deploying next-generation autonomous agents in complex environments and beyond.

\bibliography{conference}
\bibliographystyle{conference}
\appendix
\input{appendix}

\end{document}

%% file: math_commands.tex
\usepackage{amsmath,amsfonts,bm}

\def\eqref#1{equation~\ref{#1}}

\def\1{\bm{1}}

\DeclareMathAlphabet{\mathsfit}{\encodingdefault}{\sfdefault}{m}{sl}
\SetMathAlphabet{\mathsfit}{bold}{\encodingdefault}{\sfdefault}{bx}{n}



%% file: appendix.tex
\section{Atomic Operator Definition}
\label{appendix: atomic operator}

To rigorously formalize the internal mechanics of our state machine, we need to establish the semantic definitions of its foundational building blocks. 
These atomic operators dynamically manipulate different memory subspaces of the system state $\mathcal{S}$ and subsequently construct the global transition matrix:

\paragraph{Advancement Operator [$\delta: \mathcal{C} \rightarrow \mathcal{C}$].} It manages the macroscopic task progression during the milestone state. Upon the success of a milestone, $\delta$ pops the completed sub-task $c^{(0)}$ from the head of the checklist queue $c$, formally defined as:
\begin{equation}
    \delta(c) = c \setminus \{ c^{(0)} \} 
\end{equation}

\paragraph{Folding Operator [$\phi: \mathcal{A} \rightarrow \mathcal{M}$].} As a cognitive compression mechanism triggered during a milestone, it projects a high-dimensional reasoning trajectory, i.e., the active path $a$, into a condensed semantic knowledge representation:
\begin{equation}
    \phi(a) = \mathcal{F}_{\text{comp}}\left( \bigoplus_{i=1}^{|a|} \tau_i \right) 
\end{equation}
where $\mathcal{F}_{\text{comp}}$ denotes the auto-regressive abstraction function that semantically synthesizes the sequential HAR tuples. 
To preserve short-term context and ensure reasoning continuity, the folding operation selectively retains the most recent HAR micro-chains within the active memory rather than compressing the entire trajectory.

\paragraph{Extraction Operator [$\ell: \mathcal{T} \rightarrow \mathcal{P}$].}
Upon entering a failure state, the framework first initiates a distillation stage to convert raw error data into actionable knowledge. The extraction operator $\ell(\tau_t)$ auto-regressively analyzes the rejected tuple $\tau_t$ to deduce the underlying technical cause. 

Instead of merely recording the error, it generates a concise instructional synthesis besides the model-generated pitfall memory $p$. For instance, rather than appending raw traceback logs, it produces a high-level heuristic: \texttt{"Avoid repeatedly reading the same directory; pivot to test-case analysis to verify current assumptions."} 
This ensures that the essence of the failure is preserved in a format that directly informs future decision-making.

\paragraph{Update Operator [$\mu: \mathbb{W} \times \mathcal{T} \rightarrow \mathbb{W}$].}
The system will perform loop monitoring to identify and penalize repetitive, flawed behaviors. The update operator $\mu$ dynamically increments a loop weight $w_t$ based on the similarity between the current failure and historical trajectory nodes:
\begin{equation}
    \mu(w_{t-1}, \tau_t) = w_{t-1} + \gamma \cdot \mathbf{1}_{\text{sim}}(\tau_t, w_{t-1}) 
\end{equation} 
where $\gamma$ denotes the penalty increment and $\mathbf{1}_{\text{sim}}$ is an indicator function. 
When a repeat failure is detected ($\mathbf{1}_{\text{sim}} = 1$), the framework merges redundant logs into a compact count-based representation (e.g., \texttt{"Action [Action] failed $N$ times"}). 
Once $w_t$ exceeds a predefined threshold, the system triggers a structural intervention—such as dynamic prompt switching or hard gating—to forcefully steer the agent away from local minima and toward an alternative reasoning trajectory.

\paragraph{Pruning Operator [$\pi: \mathcal{A} \rightarrow \mathcal{A}$].}
The system will implement context sanitation to physically reset the reasoning environment when the model is trapped in the failure loop. 
To prevent invalid or contaminated reasoning chains from bloating the context window and misleading future steps, the pruning operator $\pi$ executes a defensive rollback on the active path sequence $a = (\tau_1, \dots, \tau_k)$:
\begin{equation}
    \pi(a) = (\tau_1, \dots, \tau_{k-\eta}) 
\end{equation}
where $\eta \ge 1$ represents the rollback depth to the last verified stable node. Unlike simple truncation, this operation is strategically coupled with the insights from $\ell$ and $\mu$; the agent resumes from a clean state but carries the newly distilled pitfall memory, ensuring that the subsequent exploration does not replicate the previously pruned logic branches.

\paragraph{Check Operator [$\chi: \mathcal{W} \rightarrow \{\texttt{SAFE}, \texttt{WARN}\}$].} 

It acts as an outer observer for routine monitoring evaluation, verifying the current reasoning trajectory against the loop alerts $w$ to proactively monitor for loop behaviors. It can be formulated as a thresholding function:
\begin{equation}
    \chi(w) = 
    \begin{cases} 
      \texttt{WARN}, & \text{if } \exists \omega \in w \text{ s.t. } \omega \ge \theta_{\text{loop}} \\
      \texttt{SAFE}, & \text{else} 
    \end{cases} 
\end{equation}
where $\theta_{\text{loop}}$ represents the maximum tolerable threshold for repetitive behaviors.

\section{Compression-Control Parameter}
\label{appendix: state_compression_ratio}

Equation~\ref{equ: window} uses the scaling factor
\begin{equation}
s(\rho)=\frac{1-\rho}{\rho},
\end{equation}
for $\rho>0$, while $\rho=0$ is treated separately as the maximum-retention setting, where the capacity is capped only by $L_{\max}$. Thus, smaller $\rho$ values preserve more historical context, whereas larger $\rho$ values impose more aggressive compression. 

At $\rho=0.5$, $s(\rho)=1$, so the corresponding capacity equals the reference capacity $V_r$ before applying the global upper bound. At $\rho=1$, $s(\rho)=0$, and each controlled component reaches its predefined minimum capacity $V_{\min}$.

More concretely, $\rho$ jointly controls the capacities of the \emph{structured system state} and the \emph{episodic raw trajectory}, while operational directives remain fixed. The controlled quantities include:

  \begin{itemize}
      \item \textbf{Recent Raw turn Window}: the number of most recent complete turns retained in raw form.
      \item \textbf{Recent Raw Assistant Window}: the number of recent assistant turns whose original action text is preserved.
      \item \textbf{Recent Raw Unclipped Window}: the number of latest turns for which observations are preserved with looser clipping.
      \item \textbf{Active Reasoning Path Window}: the maximum number of reasoning-path records retained in the active state.
      \item \textbf{Post-Milestone Active Path Window}: the tighter cap applied to the active reasoning path after a \texttt{MILESTONE} fold.
      \item \textbf{Pitfall Window}: the number of retained failure-derived lessons.
      \item \textbf{Milestone Window}: the number of verified logical milestones retained in compressed state.
      \item \textbf{Online Context Entry Budget}: the maximum number of compressed state entries exposed to the online controller.
      \item \textbf{Output Clipping Budgets}: character-level caps for different observation types, including:
      \begin{itemize}
          \item general latest output budget,
          \item test/traceback budget,
          \item diff budget,
          \item source/read/view/search budget,
          \item other output budget.
      \end{itemize}
  \end{itemize}

Each controlled quantity has its own $(V_r,V_{\min})$ configuration. Therefore, a single compression-control parameter $\rho$ consistently adjusts raw-history retention, structured-state capacity, and observation fidelity, providing a continuous trade-off between context preservation and compression.

\section{Experiments}

\subsection{Experimental Setup}
\label{appendix: exp_setup}

\paragraph{Framework Implementation and Configuration. }
To ensure reproducibility, all experimental parameters are standardized via YAML configuration files. 
For \textbf{WebShop}, we evaluate Qwen2.5-14B-Instruct over 1,000 episodes with $\text{temperature}=0$, a limit of 20 environment steps and 30 model calls per task, and at most 1,024 generated tokens per call. For 7B models, We evaluate with the same limits and 8,192 token context window. The baseline and corresponding ARSM model on Deepseek-V4.1-Flash and GLM-5.3-Flash are accessed directly through third-party APIs.
For the \textbf{16-Objective Multi-Hop QA}, we assess Qwen2.5-14B-Instruct across 300 instances under zero-temperature setting and 1,024-token budget, permitting up to 20 interaction turns alongside an E5 retriever that fetches the top-3 passages per query.
For \textbf{SWE-Bench Lite}, to accommodate lengthy reasoning traces, we expand the context horizon for Qwen3.5-9B and Qwen3.5-27B to at most 300 agent steps and 8,192 tokens per call with a sampling temperature of $0.6$.

\paragraph{Computational Resources. } Our experiments and development were conducted on two primary clusters: one equipped with eight NVIDIA H200 GPUs and another with four NVIDIA RTX A6000 GPUs. 
Most models were served through vLLM. 

\subsection{Evaluation Metrics}
\label{appendix: eval_metrics}

To evaluate the performance and efficiency of ARSM and the baseline models, we employ different metrics:

\begin{itemize}
    \item For Webshop, we measure average final reward and token usage. 
    \item For Multi-Hop QA, we measure exact matching (EM), F1, and token usage.
    \item For SWE-Bench Lite, we evaluate the models using Pass@1 and Pass@5 and token usage. 
\end{itemize}

These metrics collectively provide a general view of the model's capability, and they are defined as:

\paragraph{Token Usage.} To assess the computational cost and resource efficiency, we track the total number of tokens consumed during the entire inference process. It contains both the prompt tokens and the completion tokens. Lower token usage indicates higher inference efficiency and reduced API costs.
We also measure peak token usage, the maximum context window size reached at any single point during the model's reasoning or generation process.

\paragraph{WebShop Reward.}
The WebShop reward measures how well the purchased product satisfies the user's shopping instruction. It considers the product attributes, purchase options, price constraint, and product type:
\begin{equation}
R = r_{\mathrm{type}}
\frac{N_{\mathrm{attr}} + N_{\mathrm{option}} + I_{\mathrm{price}}}
{N_{\mathrm{attr}}^{*} + N_{\mathrm{option}}^{*} + 1},
\end{equation}
where $N_{\mathrm{attr}}$ and $N_{\mathrm{option}}$ are the numbers of matched attributes and options, respectively, and $I_{\mathrm{price}}=1$ if the product satisfies the price constraint and $0$ otherwise. The coefficient $r_{\mathrm{type}}$ measures whether the selected product has the correct product type. The final reward ranges from $[0,1]$, and $R=1$ indicates that all requirements are satisfied.
  
\paragraph{Exact Match (EM), F1 Score, and Multi-Objective Score.} 
For each objective in Multi-Objective Multi-Hop QA, Exact Match (EM) measures whether the generated prediction exactly matches the ground-truth answer, while the F1 score measures token-level overlap between the prediction and the reference answer:
\begin{equation}
\text{F1} = \frac{2 \times \text{Precision} \times \text{Recall}}
{\text{Precision} + \text{Recall}}.
\end{equation}

Following the 16-objective Multi-Hop QA setting, we evaluate each instance over 16 objectives. We report the aggregated score as the sum of the scores across all objectives:
\begin{equation}
\text{Score} = \sum_{i=1}^{16} s_i \in [0,16],
\end{equation}
where $s_i$ denotes the score for the $i$-th objective. 

\paragraph{Pass@$k$.} It measures the functional correctness of the generated solutions. We generate $n$ samples for each task ($n \ge k$) and count the number of correct samples $c$. The unbiased estimator for Pass@$k$ is defined as:
\begin{equation}
   \text{Pass@}k = \mathbb{E}_{\text{Problems}} \left[ 1 - \frac{\binom{n-c}{k}}{\binom{n}{k}} \right] 
\end{equation}
In our experiments on SWE-Bench Lite, we report the success rate using $k=1$ and $k=5$. These metrics respectively reflect the model's efficiency in surfacing internal knowledge and the actual expansion of its underlying capability frontier.

\subsection{More Experiments}
\label{appendix: more_experiment}

\begin{figure}[htbp]
    \centering
    \includegraphics[width=\linewidth]{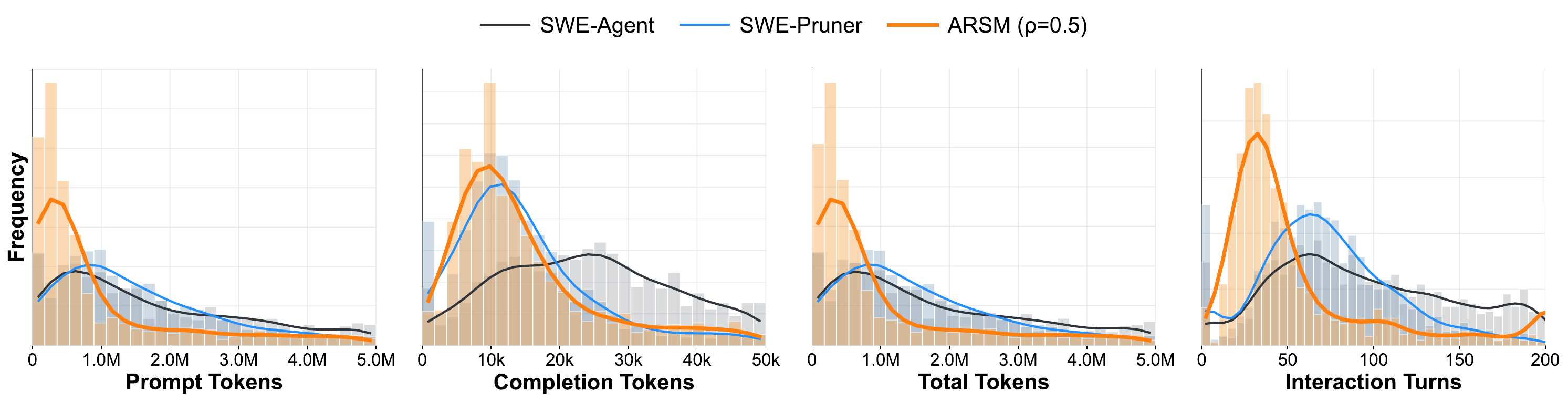}
    \caption{Efficiency analysis on SWE-Bench Lite using Qwen3.5-9B.}
    \label{fig: turn_9b}
\end{figure}

\paragraph{Efficiency Distribution on Qwen3.5-9B.} To provide a comprehensive understanding of our method's generalization ability, we present additional experimental results focusing on the smaller backbone model, Qwen3.5-9B. 
Figure~\ref{fig: turn_9b} illustrates the detailed distribution of computational costs across four dimensions: prompt tokens, completion tokens, total tokens, and interaction turns. 

Similar to the trends observed with the larger model, the kernel density estimation curves for ARSM ($\rho=0.5$) exhibit a pronounced leftward shift compared to both SWE-Agent and SWE-Pruner. 
Notably, the distributions for interaction turns and prompt tokens are highly concentrated in the lower-cost regions. This indicates that even with a weaker reasoning backbone, our state compression mechanism consistently prevents context over-accumulation and guides the agent to reach the correct patch in fewer exploration steps, thereby avoiding the long-tail inefficiency prevalent in standard autonomous agents.

However, a residual long tail in interaction turns remains. We attribute this primarily to the inherent instruction-following limitations of smaller models; specifically, the 9B model occasionally struggles to reliably generate the structured summarizations (e.g., reflections, pitfalls, and aims) required by our prompts, necessitating additional recovery turns.

\begin{figure}[htbp]
    \centering
    \includegraphics[width=\linewidth]{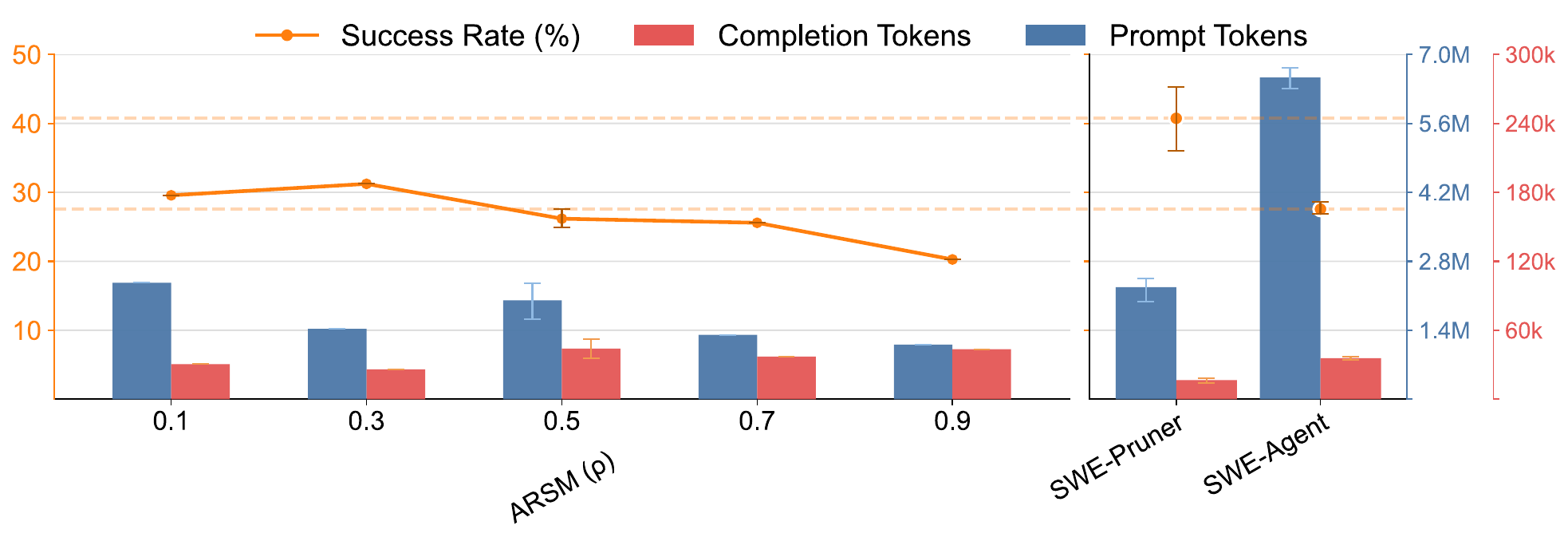}
    \caption{Performance and efficiency comparison across different compression-control parameters ($\rho$) and baselines using Qwen3.5-9B.}
    \label{fig: ablation_9b}
\end{figure}

\paragraph{Impact of Compression-Control Parameter on Qwen3.5-9B.} As shown in Figure~\ref{fig: ablation_9b}, baseline SWE-Agent consumes about 7.0M tokens, which severely limits its practical deployment. 
In stark contrast, ARSM introduces a massive reduction in prompt token usage across all evaluated compression rates, keeping the payload strictly under 3.0M tokens. 
As $\rho$ increases (indicating more aggressive compression), the prompt token cost monotonically decreases.

Crucially, this substantial cost reduction does not lead to a catastrophic collapse in performance. While extremely aggressive compression (e.g., $\rho=0.9$) inevitably discards some necessary context and causes a drop in the success rate, intermediate values successfully maintain a competitive success rate, showing that ARSM provides a highly effective and tunable trade-off space.

\begin{figure}[htbp]
    \centering
    \includegraphics[width=\linewidth]{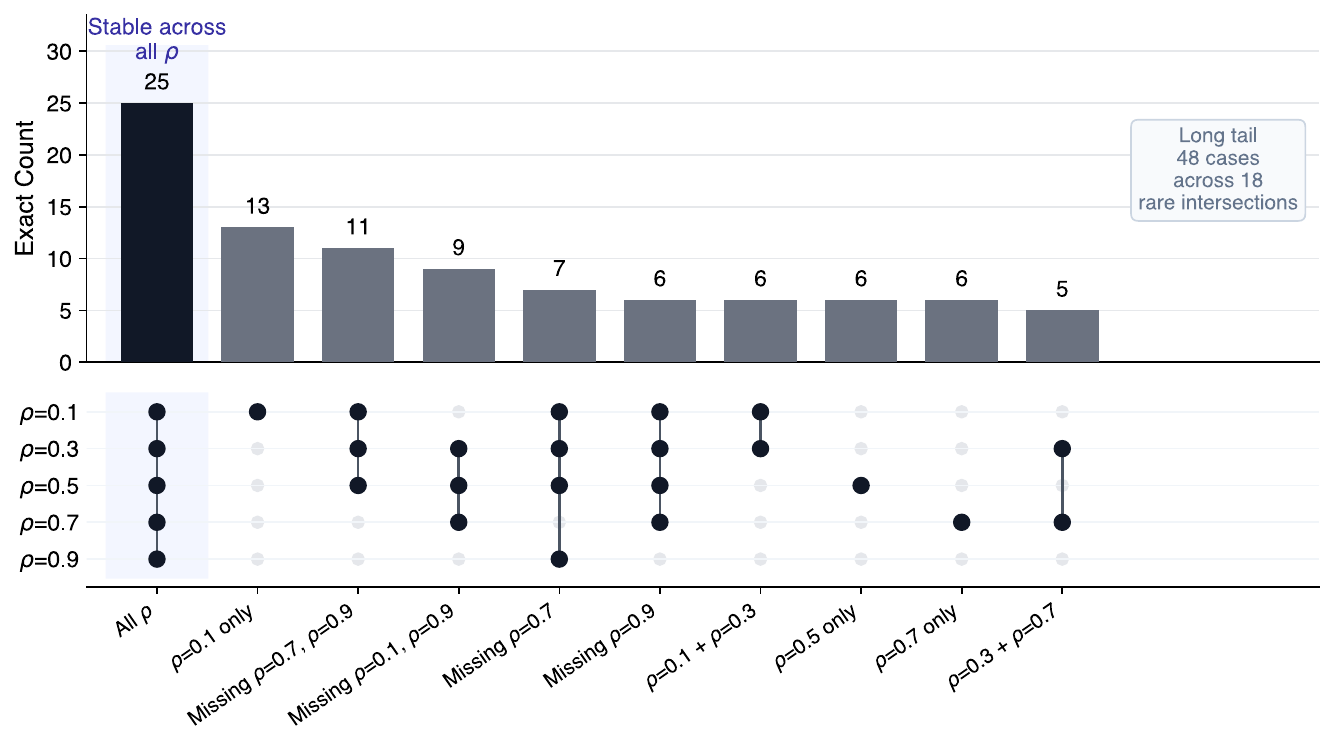}
    \caption{Intersections of resolved SWE-Bench Lite instances on Qwen3.5-9B model.}
    \label{fig: intersec_9b}
\end{figure}

\begin{figure}[htbp]
    \centering
    \includegraphics[width=\linewidth]{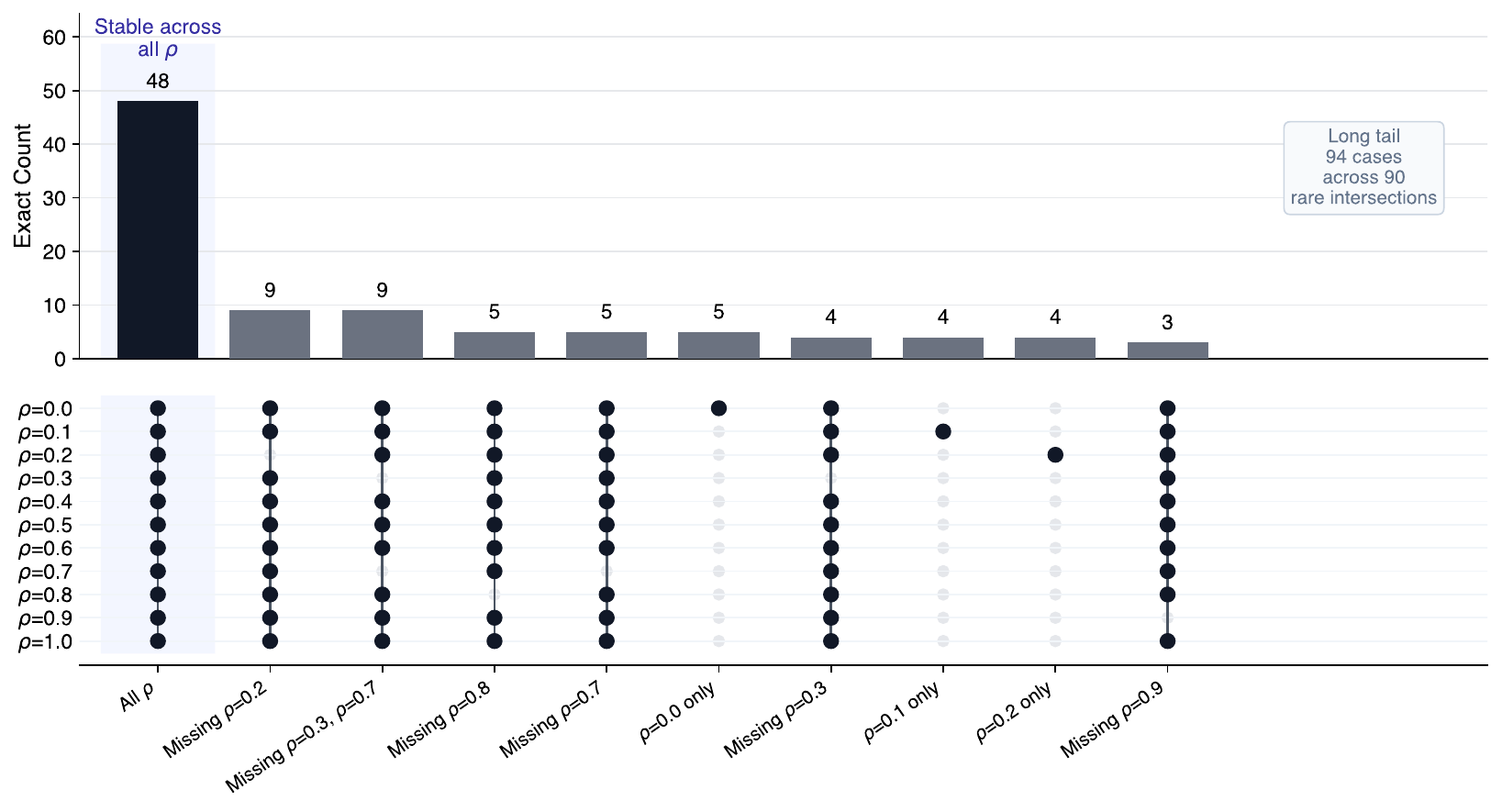}
    \caption{Intersections of resolved SWE-Bench Lite instances on Qwen3.5-27B model.}
    \label{fig: intersec_27b}
\end{figure}

\paragraph{Effect of $\rho$ on Solved-Set Overlap.} We analyze the exact overlap of instances solved by ARSM under different $\rho$ settings for both Qwen3.5-9B and Qwen3.5-27B, as shown in Figure~\ref{fig: intersec_9b}\&\ref{fig: intersec_27b}. 

Across model scales, ARSM exhibits a clear stable core: 25 instances are solved by all evaluated $\rho$ settings on Qwen3.5-9B, and 48 instances are solved by all 11 $\rho$ settings on Qwen3.5-27B. 
The largest intersections are dominated by fully stable or near-stable solutions, where only a small number of $\rho$ values fail to solve the instance. 
At the same time, both models show a non-trivial long tail of rare intersection patterns, indicating that varying $\rho$ also expands coverage by enabling ARSM to solve additional instances beyond the stable core. 

Overall, these results show that ARSM has a strong shared solving capability across $\rho$ settings, while the variation in $\rho$ further broadens its coverage through complementary exploration.

\begin{table}[htbp]
    \centering
    \caption{Solved-set overlap of ARSM across different $\rho$ settings.}
    \begin{tabular}{lcccc}
        \toprule
        Model & Union Solved & Solved by All $\rho$ & Long-Tail Coverage \\
        \midrule
        Qwen3.5-9B  & 142 & 25 & 48 cases / 18 patterns \\
        Qwen3.5-27B & 190 & 48 & 94 cases / 90 patterns \\
        \bottomrule
        \end{tabular}
\end{table}

\subsection{Experiment Protocol}
\label{appendix: exp_protocol}

\paragraph{Evaluation Protocol of Trajectory Summarization and State Machine Evolution.}
We evaluate five variants to examine the effects of trajectory summarization and state machine evolution:
\begin{itemize}
    \item \textbf{Plain Summary} repeatedly rewrites the interaction history as a single free-form summary.
    \item \textbf{Structured Summary} rewrites a single summary with fixed fields for the goal, progress, important observations, failures, and next action.
    \item \textbf{HAR-only} accumulates hypothesis--action--result records in chronological order. When the context budget is reached, it removes
      the oldest complete records.
    \item \textbf{SM-only} retains the dynamic state machine, including its checklist, candidate ledger, pitfalls, milestones, active path, routing signals, and state transitions. It uses generic structured turn summaries instead of HAR records.
    \item \textbf{Full ARSM} combines HAR records with the dynamic state machine.
\end{itemize}

\paragraph{Trajectory-Level Evaluation Protocol.}
To evaluate the quality of ARSM's compressed state and routing decisions, we conduct a trajectory-level analysis on 100 WebShop trajectories generated with $\rho=0.5$. We use DeepSeek-V4.1-Flash~\citep{deepseekai2026deepseekv41flash} as the LLM-based evaluator. For each trajectory, the evaluator assesses every interaction turn independently under four criteria: state fidelity, state sufficiency, routing correctness, and trajectory coherence,
\begin{equation} 
s_{j,t}^{(m)} \in \{0,1\}, 
\end{equation}
where $j$ indexes the trajectory, $t$ indexes the interaction turn, and $m$ denotes the evaluation metric. A score of $1$ indicates that the turn satisfies the corresponding criterion, while $0$ indicates otherwise. We first compute the average score within each trajectory: \begin{equation} \bar{s}_{j}^{(m)} = \frac{1}{T_j} \sum_{t=1}^{T_j} s_{j,t}^{(m)}, \end{equation} where $T_j$ is the number of turns in trajectory $j$. We then report the final metric by averaging the trajectory-level scores over all 100 trajectories: \begin{equation} S^{(m)} = \frac{1}{100} \sum_{j=1}^{100} \bar{s}_{j}^{(m)}. \end{equation} The resulting value is reported as a percentage in Table~\ref{tab:webshop_state_quality}. This two-stage aggregation gives equal weight to each trajectory, rather than allowing longer trajectories with more turns to dominate the final score.

\section{Illustration of State Representation}
\label{appendix:state_representation}

We illustrate what ARSM actually stores in its two main memory surfaces.
Using the instance \texttt{django\_\_django-16408} as a running example, we show how local reasoning traces are represented in (i) recent raw interaction memory, and (ii) compressed internal state.

\subsection{Recent Raw Interaction Memory}

Recent raw memory preserves a bounded portion of the original interaction trajectory in a near-transcript form. Each preserved segment typically contains four elements: Hypothesis, Action, Observation and Result.

In our implementation, this memory surface is formatted as a lightweight HAR chain record.
A representative example from \texttt{django\_\_django-16408} ($\rho=0.5$, Turn 20) is shown below:

\begin{promptmint}{ARSM Trajectory Example}
	\begin{lstlisting}[basicstyle=\ttfamily\small, breaklines=true, columns=fullflexible]
<conclusion>
[REFLECTION] Where: Reading query_utils.py Result: The output was truncated, showing only line numbers and summaries. The test failure shows that when using FilteredRelation with select_related, the wrong related object is returned (PoolStyle instead of Tournament). Conclusion: Need to see the full FilteredRelation class implementation to understand the bug.
[STATUS] ONGOING
[PITFALL] N/A
</conclusion>
<reasoning type=\"CONTINUE\">
[CURRENT_STEP] Understand current logic
[AIM] Read the full FilteredRelation class implementation from query_utils.py
</reasoning>
<mswea_bash_command>
\ngrep -n \"class FilteredRelation\" /testbed/django/db/models/query_utils.py
</mswea_bash_command>
	\end{lstlisting}
\end{promptmint}

This representation is intentionally close to the original interaction trace. 
It is useful for preserving local procedural detail, especially when the model may still need to reuse exact command outcomes, recently opened file fragments, or immediate debugging evidence.

\subsection{Compressed Internal State}

In contrast, the compressed internal state is not a transcript replay.
Instead, it reorganizes the trajectory into a structured summary that retains reasoning continuity while discarding low-value interaction redundancy.

A representative internal-state excerpt for \texttt{django\_\_django-16408} ($\rho=0.5$, Turn 50) is shown below:

\begin{promptmint}{ARSM Internal State Example}
	\begin{lstlisting}[basicstyle=\ttfamily\small, breaklines=true, columns=fullflexible]
[Internal State Update]
# Global Checklist
[x] 1. Exploring the codebase structure to understand the issue
[ ] 2. Understand current logic
[ ] 3. Identify root cause
[ ] 4. Patch the source
[ ] 5. Verify / submit

# Pitfall
- Need to use Django's test infrastructure instead of direct Python imports.
- Django test runner location needs to be discovered.
- Django needs to be installed in development mode from the testbed directory before running tests.
...

# Milestones
- No verified milestones yet.

# Loop Alerts
- Inspect loop warning: the current step is repeatedly reading the same file/path set 3 times without new evidence. Stop inspecting this target; if a patch exists inspect `git diff -- .` and submit with `echo COMPLETE_TASK_AND_SUBMIT_FINAL_OUTPUT`, otherwise edit the target or mark FAILURE and pivot.
- Execution guard active: repeated looping has blocked this command path. You must change the step, file target, or action type.
...

# Active Reasoning Path
- Turn 1: [Hypothesis] Find the relevant files for FilteredRelation and the test case [Action] find /testbed -type f -name \"*.py\" | xargs grep -l \"FilteredRelation\" | head -20 </mscontent> [Reflection] /testbed/django/db/models/query.py /testbed/django/db/models/query_utils.py /testbed/django/db/models/sql/query.py /testbed/django/db/models/__init__.py /testbed/tests/filtered_relation/tests.py /testbed/tests/select_related_onetoone/tests.py /testbed/tests/queryset_pickle/tests.py /testbed/tests/known_related_objects/tests.py bash: -c: line 2: syntax error\u2026. Found relevant files containing FilteredRelation. Need to examine the test file to understand the issue context.
- Turn 2: [Hypothesis] Read the known_related_objects/tests.py file to understand the test and models [Action] cat /testbed/tests/known_related_objects/tests.py [Reflection] from django.db.models import FilteredRelation from django.test import TestCase from .models import Organiser, Pool, PoolStyle, Tournament class ExistingRelatedInstancesTests(TestCase): @classmethod def setUpTestData(cls): cls.t1 = Tournament.objects.create(name=\"Tourney 1\") cls.t2 = Tournament.objects.create(name=\"Tourney 2\") cls.o1 = Organiser.objects.crea\u2026. Found the test file. Need to see the full content including the test case that's failing and the models.
...
	\end{lstlisting}
\end{promptmint}

Compared with recent raw memory, this format is more abstract. It no longer stores each interaction as a full transcript fragment. Instead, it preserves the causal reasoning trace in a compressed form that is easier to carry across long horizons.

\subsection{Interpretation}

The distinction between these two memory surfaces is central to ARSM.

Recent raw memory preserves \emph{interaction fidelity}: it keeps the model close to the original tool-use trace and therefore supports short-horizon procedural reuse.
Compressed internal state preserves \emph{reasoning continuity}: it extracts stable conclusions, reusable pitfall lessons, and active hypotheses from that trace, making them available even when the original interaction history has been pruned.

This distinction between structured system state and episodic raw trajectory becomes especially important when the compression-control parameter $\rho$ changes. As $\rho$ increases, ARSM does not merely reduce the amount of remembered history; it shifts a larger fraction of task-relevant continuity from transcript-like raw replay into structured internal state.
This effect is examined next in our trajectory comparison case study.

\section{Case Study: Trajectory Comparison Under Different Compression-Control Parameter}
\label{appendix:rho_case_study}

To provide a concrete illustration of how ARSM changes the organization of reasoning history under different compression settings, we present an instance-level case study on
\texttt{django\_\_django-16408}.
We compare the same task under three compression-control parameters: $\rho=0.2$, $\rho=0.5$, and $\rho=0.8$.

\subsection{Trajectory Statistics}

Table~\ref{tab: django16408_overview} summarizes the overall trajectory lengths.
The same task produces substantially different rollout lengths under different compression regimes:
the moderate setting $\rho=0.5$ leads to the longest trajectory, while $\rho=0.2$ and $\rho=0.8$ terminate earlier.

\begin{table}[htbp]
	\centering
	\caption{Trajectory length comparison under different compression-control parameters.}
	\begin{tabular}{cccc}
		\toprule
		\textbf{Instance} & \textbf{$\rho=0.2$} & \textbf{$\rho=0.5$} & \textbf{$\rho=0.8$} \\
		\midrule
		\texttt{django\_\_django-16408} & 151 & 300 & 207 \\
		\bottomrule
	\end{tabular}
	\label{tab: django16408_overview}
\end{table}

This difference indicates that $\rho$ is not merely controlling how much history is retained.
Instead, it changes the agent's effective reasoning process by shifting the balance between raw interaction replay and structured internal-state accumulation.

\subsection{Turn-Level Memory Comparison}

To make this shift explicit, we compare several aligned turns from the same instance. For each turn, we focus on two observable quantities:
(i) the amount of retained recent raw interaction context, and
(ii) the amount of information folded into the internal state update.

Qualitatively, the three settings exhibit distinct patterns:

\begin{itemize}
	\item At $\rho=0.2$, the agent keeps a larger amount of recent raw interaction history available, and the internal state grows more conservatively.
	\item At $\rho=0.5$, the controller more aggressively folds history into structured state, yielding a much larger internal summary.
	\item At $\rho=0.8$, the raw interaction surface is more selective still, and the agent relies more heavily on compact but persistent summarized state.
\end{itemize}

This behavior is consistent with the intended role of ARSM: as $\rho$ increases, context preservation shifts from transcript-level replay toward state-level continuity.

\subsection{Representative Snippets}

We next compare representative excerpts from the trajectories.
The purpose is not to compare the exact semantic correctness of each intermediate reasoning step, but rather to show how the same underlying task history is represented differently across compression regimes.

\paragraph{Structured System State.}

The strongest contrast appears in the internal state itself.
At $\rho=0.2$, the state accumulates lessons, but the representation remains relatively restrained.
At $\rho=0.5$, the state becomes much larger and encodes a long sequence of refined hypotheses about how join reuse,
nested relation resolution, and FilteredRelation paths should be handled.
At $\rho=0.8$, the state is still substantial, but comparatively more compressed than the $\rho=0.5$ case.

\paragraph{Episodic Raw Trajectory.}
In lower-compression settings, recent raw memory retains more direct traces of exploratory interaction.
At higher compression levels, the retained raw content becomes more selective and increasingly focused on the active subproblem.

\begin{table}[htbp]
	\centering
	\caption{
		\textbf{Qualitative comparison of recent raw memory. }As $\rho$ increases, the raw interaction surface becomes more selective and more localized to the current subproblem.
	}
	\begin{tabular}{p{0.31\textwidth} p{0.31\textwidth} p{0.31\textwidth}}
		\toprule
		\textbf{$\rho=0.2$} & \textbf{$\rho=0.5$} & \textbf{$\rho=0.8$} \\
		\midrule
		\ttfamily
		Recent raw interaction remains broad and preserves more direct traces of earlier file-reading failures, exploratory
		inspection, and repeated attempts to understand the relation structure around \texttt{FilteredRelation}.
		&
		\ttfamily
		Recent raw interaction is still available, but it is more selective; repeated low-value inspection is progressively
		displaced by folded state, and the retained raw messages are increasingly concentrated on specific subproblems such as
		join reuse and multi-level relation resolution.
		&
		\ttfamily
		Recent raw interaction is narrower still; the visible raw context is more sharply localized to the current debugging
		hypothesis, while earlier exploration is no longer exposed directly and instead survives primarily through structured
		memory.
		\\
		\bottomrule
	\end{tabular}

	\label{tab:django16408_raw_memory}
\end{table}

\begin{table}[htbp]
	\centering
	\caption{
		\textbf{Representative internal-state excerpts.}
		As compression increases, more of the long-range reasoning trace is preserved through structured pitfalls instead of direct transcript replay.
	}
	\begin{tabular}{p{0.31\textwidth} p{0.31\textwidth} p{0.31\textwidth}}
		\toprule
		\textbf{$\rho=0.2$} & \textbf{$\rho=0.5$} & \textbf{$\rho=0.8$} \\
		\midrule
		\ttfamily
		\# Pitfall \newline
		- The file-reading commands are producing malformed output; need a more reliable reading method. \newline
		- Need to use grep or targeted inspection to locate FilteredRelation first. \newline
		- The problem likely concerns how filtered relation paths are resolved in the ORM.
		&
		\ttfamily
		\# Pitfall \newline
		- The fix needs to be in setup\_joins so that existing joins are reused when the field is already part of a FilteredRelation path. \newline
		- The single-level fix is insufficient; nested paths such as tournament\_pool. tournament require the second-level relation to use the correct join. \newline
		- The necessary information may need to be propagated through the relation metadata structure rather than patched only at the surface.
		&
		\ttfamily
		\# Pitfall \newline
		- The current fixes address part of the join path but still cache the intermediate object on the wrong model. \newline
		- The issue appears deeper in how nested filtered relations are represented in the relation chain. \newline
		- Submission may need to avoid unrelated test edits and focus on source-level correction only.
		\\
		\bottomrule
	\end{tabular}
	\label{tab:django16408_internal_state}
\end{table}

\subsection{Interpretation}

This case study illustrates the intended operational effect of ARSM.
Increasing $\rho$ does not simply delete history.
Instead, it changes \emph{where} the task-relevant information lives.

Under smaller $\rho$, the model can continue to rely on a larger volume of recent raw interaction history.
Under larger $\rho$, the controller increasingly shifts continuity into explicit state variables, such as lessons, milestones, and active reasoning-path summaries.
In other words, ARSM converts history from a transcript-centric representation into a state-centric representation.

The \texttt{django\_\_django-16408} example is useful precisely because it shows this transition in a nontrivial long-horizon setting.
The three trajectories do not merely differ in length; they differ in memory substrate.
This provides direct qualitative evidence that the compression-control parameter controls not only context quantity, but also the form in which reasoning continuity is preserved.

%% file: conference.bib
@article{wei2022chain,
  title={Chain-of-thought prompting elicits reasoning in large language models},
  author={Wei, Jason and Wang, Xuezhi and Schuurmans, Dale and Bosma, Maarten and Xia, Fei and Chi, Ed and Le, Quoc V and Zhou, Denny and others},
  journal={Advances in neural information processing systems},
  volume={35},
  pages={24824--24837},
  year={2022}
}

@article{schick2023toolformer,
  title={Toolformer: Language models can teach themselves to use tools},
  author={Schick, Timo and Dwivedi-Yu, Jane and Dess{\`\i}, Roberto and Raileanu, Roberta and Lomeli, Maria and Hambro, Eric and Zettlemoyer, Luke and Cancedda, Nicola and Scialom, Thomas},
  journal={Advances in neural information processing systems},
  volume={36},
  pages={68539--68551},
  year={2023}
}

@inproceedings{yao2023react,
  title = {{ReAct}: Synergizing Reasoning and Acting in Language Models},
  author = {Yao, Shunyu and Zhao, Jeffrey and Yu, Dian and Du, Nan and Shafran, Izhak and Narasimhan, Karthik and Cao, Yuan},
  booktitle = {International Conference on Learning Representations (ICLR) },
  year = {2023},
  html = {https://arxiv.org/abs/2210.03629},
}

@article{shinn2023reflexion,
  title={Reflexion: Language agents with verbal reinforcement learning},
  author={Shinn, Noah and Cassano, Federico and Gopinath, Ashwin and Narasimhan, Karthik and Yao, Shunyu},
  journal={Advances in neural information processing systems},
  volume={36},
  pages={8634--8652},
  year={2023}
}

@article{yang2024swe,
  title={Swe-agent: Agent-computer interfaces enable automated software engineering},
  author={Yang, John and Jimenez, Carlos E and Wettig, Alexander and Lieret, Kilian and Yao, Shunyu and Narasimhan, Karthik and Press, Ofir},
  journal={Advances in Neural Information Processing Systems},
  volume={37},
  pages={50528--50652},
  year={2024}
}

@inproceedings{guo2025se,
  title={SE-Agent: Self-Evolution Trajectory Optimization in Multi-Step Reasoning with LLM-Based Agents},
  author={Guo, Yifu and Lin, Jiaye and Wang, Huacan and Han, Yuzhen and Hu, Sen and Ni, Ziyi and Wang, Licheng and Chen, Mingguang},
  booktitle={The Thirty-ninth Annual Conference on Neural Information Processing Systems},
  year={2025}
}

@inproceedings{park2023generative,
  title={Generative agents: Interactive simulacra of human behavior},
  author={Park, Joon Sung and O'Brien, Joseph and Cai, Carrie Jun and Morris, Meredith Ringel and Liang, Percy and Bernstein, Michael S},
  booktitle={Proceedings of the 36th annual acm symposium on user interface software and technology},
  pages={1--22},
  year={2023}
}

@inproceedings{zhong2024memorybank,
  title={Memorybank: Enhancing large language models with long-term memory},
  author={Zhong, Wanjun and Guo, Lianghong and Gao, Qiqi and Ye, He and Wang, Yanlin},
  booktitle={Proceedings of the AAAI conference on artificial intelligence},
  year={2024}
}

@article{packer2023memgpt,
	title={Memgpt: Towards llms as operating systems},
	author={Packer, Charles and Wooders, Sarah and Lin, Kevin and Fang, Vivian and Patil, Shishir G and Stoica, Ion and Gonzalez, Joseph E},
	journal={arXiv preprint arXiv:2310.08560},
	year={2023}
}

@article{wang2023augmenting,
  title={Augmenting language models with long-term memory},
  author={Wang, Weizhi and Dong, Li and Cheng, Hao and Liu, Xiaodong and Yan, Xifeng and Gao, Jianfeng and Wei, Furu},
  journal={Advances in Neural Information Processing Systems},
  volume={36},
  pages={74530--74543},
  year={2023}
}

@article{fang2025attentionrag,
  title={Attentionrag: Attention-guided context pruning in retrieval-augmented generation},
  author={Fang, Yixiong and Sun, Tianran and Shi, Yuling and Gu, Xiaodong},
  journal={arXiv preprint arXiv:2503.10720},
  year={2025}
}

@article{kang2025acon,
  title={Acon: Optimizing context compression for long-horizon llm agents},
  author={Kang, Minki and Chen, Wei-Ning and Han, Dongge and Inan, Huseyin A and Wutschitz, Lukas and Chen, Yanzhi and Sim, Robert and Rajmohan, Saravan},
  journal={arXiv preprint arXiv:2510.00615},
  year={2025}
}

@inproceedings{jiang2023llmlingua,
  title={Llmlingua: Compressing prompts for accelerated inference of large language models},
  author={Jiang, Huiqiang and Wu, Qianhui and Lin, Chin-Yew and Yang, Yuqing and Qiu, Lili},
  booktitle={Proceedings of the 2023 conference on empirical methods in natural language processing},
  pages={13358--13376},
  year={2023}
}

@inproceedings{li2023compressing,
  title={Compressing context to enhance inference efficiency of large language models},
  author={Li, Yucheng and Dong, Bo and Guerin, Frank and Lin, Chenghua},
  booktitle={Proceedings of the 2023 conference on empirical methods in natural language processing},
  pages={6342--6353},
  year={2023}
}

@inproceedings{
    jimenez2024swebench,
    title={{SWE}-bench: Can Language Models Resolve Real-world Github Issues?},
    author={Carlos E Jimenez and John Yang and Alexander Wettig and Shunyu Yao and Kexin Pei and Ofir Press and Karthik R Narasimhan},
    booktitle={The Twelfth International Conference on Learning Representations},
    year={2024},
    url={https://openreview.net/forum?id=VTF8yNQM66}
}

@article{lewis2020retrieval,
  title={Retrieval-augmented generation for knowledge-intensive nlp tasks},
  author={Lewis, Patrick and Perez, Ethan and Piktus, Aleksandra and Petroni, Fabio and Karpukhin, Vladimir and Goyal, Naman and K{\"u}ttler, Heinrich and Lewis, Mike and Yih, Wen-tau and Rockt{\"a}schel, Tim and others},
  journal={Advances in neural information processing systems},
  volume={33},
  pages={9459--9474},
  year={2020}
}

@article{su2026u,
  title={U-Fold: Dynamic Intent-Aware Context Folding for User-Centric Agents},
  author={Su, Jin and Fang, Runnan and Li, Yeqiu and Wang, Xiaobin and Cai, Shihao and Xie, Pengjun and Zhang, Ningyu and Yuan, Fajie},
  journal={arXiv preprint arXiv:2601.18285},
  year={2026}
}

@article{ye2025agentfold,
  title={AgentFold: Long-Horizon Web Agents with Proactive Context Management},
  author={Ye, Rui and Zhang, Zhongwang and Li, Kuan and Yin, Huifeng and Tao, Zhengwei and Zhao, Yida and Su, Liangcai and Zhang, Liwen and Qiao, Zile and Wang, Xinyu and others},
  journal={arXiv preprint arXiv:2510.24699},
  year={2025}
}

@article{li2023unlocking,
  title={Unlocking context constraints of llms: Enhancing context efficiency of llms with self-information-based content filtering},
  author={Li, Yucheng},
  journal={arXiv preprint arXiv:2304.12102},
  year={2023}
}

@article{wang2026swe,
  title={SWE-Pruner: Self-Adaptive Context Pruning for Coding Agents},
  author={Wang, Yuhang and Shi, Yuling and Yang, Mo and Zhang, Rongrui and He, Shilin and Lian, Heng and Chen, Yuting and Ye, Siyu and Cai, Kai and Gu, Xiaodong},
  journal={arXiv preprint arXiv:2601.16746},
  year={2026}
}

@inproceedings{luo2026storage,
	title={From storage to experience: A survey on the evolution of llm agent memory mechanisms},
	author={Luo, Jinghao and Tian, Yuchen and Cao, Chuxue and Luo, Ziyang and Lin, Hongzhan and Li, Kaixin and Kong, Chuyi and Yang, Ruichao and Ma, Jing},
	booktitle={Findings of the Association for Computational Linguistics: ACL 2026},
	pages={41622--41652},
	year={2026}
}

@misc{qwen3.5,
    title  = {{Qwen3.5}: Towards Native Multimodal Agents},
    author = {{Qwen Team}},
    month  = {February},
    year   = {2026},
    url    = {https://qwen.ai/blog?id=qwen3.5}
}

@inproceedings{yao2022webshop,
	bibtex_show = {true},
	title = {WebShop: Towards Scalable Real-World Web Interaction with Grounded Language Agents},
	author = {Yao, Shunyu and Chen, Howard and Yang, John and Narasimhan, Karthik},
	booktitle = {ArXiv},
	year = {2022},
	html = {https://arxiv.org/abs/2207.01206},
	tag = {NLP}
}

@misc{yang2018hotpotqadatasetdiverseexplainable,
	title={HotpotQA: A Dataset for Diverse, Explainable Multi-hop Question Answering}, 
	author={Zhilin Yang and Peng Qi and Saizheng Zhang and Yoshua Bengio and William W. Cohen and Ruslan Salakhutdinov and Christopher D. Manning},
	year={2018},
	eprint={1809.09600},
	archivePrefix={arXiv},
	primaryClass={cs.CL},
	url={https://arxiv.org/abs/1809.09600}, 
}

@inproceedings{xanh2020_2wikimultihop,
	title = "Constructing A Multi-hop {QA} Dataset for Comprehensive Evaluation of Reasoning Steps",
	author = "Ho, Xanh  and
	Duong Nguyen, Anh-Khoa  and
	Sugawara, Saku  and
	Aizawa, Akiko",
	booktitle = "Proceedings of the 28th International Conference on Computational Linguistics",
	month = dec,
	year = "2020",
	address = "Barcelona, Spain (Online)",
	publisher = "International Committee on Computational Linguistics",
	url = "https://www.aclweb.org/anthology/2020.coling-main.580",
	pages = "6609--6625",
}

@article{trivedi2021musique,
	title={{M}u{S}i{Q}ue: Multihop Questions via Single-hop Question Composition},
	author={Trivedi, Harsh and Balasubramanian, Niranjan and Khot, Tushar and Sabharwal, Ashish},
	journal={Transactions of the Association for Computational Linguistics},
	year={2022},
	publisher={MIT Press}
}

@misc{aksitov2023restmeetsreactselfimprovement,
	title={ReST meets ReAct: Self-Improvement for Multi-Step Reasoning LLM Agent}, 
	author={Renat Aksitov and Sobhan Miryoosefi and Zonglin Li and Daliang Li and Sheila Babayan and Kavya Kopparapu and Zachary Fisher and Ruiqi Guo and Sushant Prakash and Pranesh Srinivasan and Manzil Zaheer and Felix Yu and Sanjiv Kumar},
	year={2023},
	eprint={2312.10003},
	archivePrefix={arXiv},
	primaryClass={cs.CL},
	url={https://arxiv.org/abs/2312.10003}, 
}

@misc{qwen2.5,
	title = {Qwen2.5: A Party of Foundation Models},
	url = {https://qwenlm.github.io/blog/qwen2.5/},
	author = {Qwen Team},
	month = {September},
	year = {2024}
}

@article{qwen2,
	title={Qwen2 Technical Report}, 
	author={An Yang and Baosong Yang and Binyuan Hui and Bo Zheng and Bowen Yu and Chang Zhou and Chengpeng Li and Chengyuan Li and Dayiheng Liu and Fei Huang and Guanting Dong and Haoran Wei and Huan Lin and Jialong Tang and Jialin Wang and Jian Yang and Jianhong Tu and Jianwei Zhang and Jianxin Ma and Jin Xu and Jingren Zhou and Jinze Bai and Jinzheng He and Junyang Lin and Kai Dang and Keming Lu and Keqin Chen and Kexin Yang and Mei Li and Mingfeng Xue and Na Ni and Pei Zhang and Peng Wang and Ru Peng and Rui Men and Ruize Gao and Runji Lin and Shijie Wang and Shuai Bai and Sinan Tan and Tianhang Zhu and Tianhao Li and Tianyu Liu and Wenbin Ge and Xiaodong Deng and Xiaohuan Zhou and Xingzhang Ren and Xinyu Zhang and Xipin Wei and Xuancheng Ren and Yang Fan and Yang Yao and Yichang Zhang and Yu Wan and Yunfei Chu and Yuqiong Liu and Zeyu Cui and Zhenru Zhang and Zhihao Fan},
	journal={arXiv preprint arXiv:2407.10671},
	year={2024}
}

@article{chen2024agent,
    title={Agent-FLAN: Designing Data and Methods of Effective Agent Tuning for Large Language Models},
    author={Chen, Zehui and Liu, Kuikun and Wang, Qiuchen and Zhang, Wenwei and Liu, Jiangning and Lin, Dahua and Chen, Kai and Zhao, Feng},
    journal={arXiv preprint arXiv:2403.12881},
    year={2024}
}

@misc{zhou2025mem1,
  title        = {MEM1: Learning to Synergize Memory and Reasoning for Efficient Long-Horizon Agents},
  author       = {Zhou, Zijian and Qu, Ao and Wu, Zhaoxuan and Kim, Sunghwan and Prakash, Alok and Rus, Daniela and Zhao, Jinhua and Low, Bryan Kian Hsiang and Liang, Paul Pu},
  year         = {2025},
  archivePrefix= {arXiv},
  primaryClass = {cs.CL},
  url          = {https://arxiv.org/abs/2506.15841},
}

@article{jin2025search,
  title={Search-r1: Training llms to reason and leverage search engines with reinforcement learning},
  author={Jin, Bowen and Zeng, Hansi and Yue, Zhenrui and Yoon, Jinsung and Arik, Sercan and Wang, Dong and Zamani, Hamed and Han, Jiawei},
  journal={arXiv preprint arXiv:2503.09516},
  year={2025}
}

@misc{wu2026contextbudgetbudgetawarecontextmanagement,
      title={ContextBudget: Budget-Aware Context Management for Long-Horizon Search Agents}, 
      author={Yong Wu and YanZhao Zheng and TianZe Xu and ZhenTao Zhang and YuanQiang Yu and JiHuai Zhu and Chao Ma and BinBin Lin and BaoHua Dong and HangCheng Zhu and RuoHui Huang and Gang Yu},
      year={2026},
      eprint={2604.01664},
      archivePrefix={arXiv},
      primaryClass={cs.AI},
      url={https://arxiv.org/abs/2604.01664}, 
}

@misc{deepseekai2026deepseekv41flash,
      title={DeepSeek-V4.1-Flash: Pushing the Limits of KV Cache Compression},
      author={DeepSeek-AI},
      year={2026},
}

@misc{glm5team2026glm5vibecodingagentic,
      title={GLM-5: from Vibe Coding to Agentic Engineering},
      author={GLM-5-Team and : and Aohan Zeng and Xin Lv and Zhenyu Hou and Zhengxiao Du and Qinkai Zheng and Bin Chen and Da Yin and Chendi Ge and Chenghua Huang and Chengxing Xie and Chenzheng Zhu and Congfeng Yin and Cunxiang Wang and Gengzheng Pan and Hao Zeng and Haoke Zhang and Haoran Wang and Huilong Chen and Jiajie Zhang and Jian Jiao and Jiaqi Guo and Jingsen Wang and Jingzhao Du and Jinzhu Wu and Kedong Wang and Lei Li and Lin Fan and Lucen Zhong and Mingdao Liu and Mingming Zhao and Pengfan Du and Qian Dong and Rui Lu and Shuang-Li and Shulin Cao and Song Liu and Ting Jiang and Xiaodong Chen and Xiaohan Zhang and Xuancheng Huang and Xuezhen Dong and Yabo Xu and Yao Wei and Yifan An and Yilin Niu and Yitong Zhu and Yuanhao Wen and Yukuo Cen and Yushi Bai and Zhongpei Qiao and Zihan Wang and Zikang Wang and Zilin Zhu and Ziqiang Liu and Zixuan Li and Bojie Wang and Bosi Wen and Can Huang and Changpeng Cai and Chao Yu and Chen Li and Chengwei Hu and Chenhui Zhang and Dan Zhang and Daoyan Lin and Dayong Yang and Di Wang and Ding Ai and Erle Zhu and Fangzhou Yi and Feiyu Chen and Guohong Wen and Hailong Sun and Haisha Zhao and Haiyi Hu and Hanchen Zhang and Hanrui Liu and Hanyu Zhang and Hao Peng and Hao Tai and Haobo Zhang and He Liu and Hongwei Wang and Hongxi Yan and Hongyu Ge and Huan Liu and Huanpeng Chu and Jia'ni Zhao and Jiachen Wang and Jiajing Zhao and Jiamin Ren and Jiapeng Wang and Jiaxin Zhang and Jiayi Gui and Jiayue Zhao and Jijie Li and Jing An and Jing Li and Jingwei Yuan and Jinhua Du and Jinxin Liu and Junkai Zhi and Junwen Duan and Kaiyue Zhou and Kangjian Wei and Ke Wang and Keyun Luo and Laiqiang Zhang and Leigang Sha and Liang Xu and Lindong Wu and Lintao Ding and Lu Chen and Minghao Li and Nianyi Lin and Pan Ta and Qiang Zou and Rongjun Song and Ruiqi Yang and Shangqing Tu and Shangtong Yang and Shaoxiang Wu and Shengyan Zhang and Shijie Li and Shuang Li and Shuyi Fan and Wei Qin and Wei Tian and Weining Zhang and Wenbo Yu and Wenjie Liang and Xiang Kuang and Xiangmeng Cheng and Xiangyang Li and Xiaoquan Yan and Xiaowei Hu and Xiaoying Ling and Xing Fan and Xingye Xia and Xinyuan Zhang and Xinze Zhang and Xirui Pan and Xu Zou and Xunkai Zhang and Yadi Liu and Yandong Wu and Yanfu Li and Yidong Wang and Yifan Zhu and Yijun Tan and Yilin Zhou and Yiming Pan and Ying Zhang and Yinpei Su and Yipeng Geng and Yong Yan and Yonglin Tan and Yuean Bi and Yuhan Shen and Yuhao Yang and Yujiang Li and Yunan Liu and Yunqing Wang and Yuntao Li and Yurong Wu and Yutao Zhang and Yuxi Duan and Yuxuan Zhang and Zezhen Liu and Zhengtao Jiang and Zhenhe Yan and Zheyu Zhang and Zhixiang Wei and Zhuo Chen and Zhuoer Feng and Zijun Yao and Ziwei Chai and Ziyuan Wang and Zuzhou Zhang and Bin Xu and Minlie Huang and Hongning Wang and Juanzi Li and Yuxiao Dong and Jie Tang},
      year={2026},
      eprint={2602.15763},
      archivePrefix={arXiv},
      primaryClass={cs.LG},
      url={https://arxiv.org/abs/2602.15763},
}
